\documentclass{article}

\usepackage{microtype}
\usepackage{graphicx}
\usepackage{subfigure}
\usepackage{booktabs} 

\usepackage{hyperref}

\usepackage[accepted]{icml2024}

\usepackage{amssymb}
\usepackage{mathtools}
\usepackage{amsthm}
\usepackage{amsfonts}

\usepackage{url}
\usepackage{epsfig}
\usepackage{bbm}

\usepackage[capitalize,noabbrev]{cleveref}

\theoremstyle{plain}

\theoremstyle{definition}

\theoremstyle{remark}

\usepackage[textsize=tiny]{todonotes}

\icmltitlerunning{Benchmarking MOEAs for solving continuous multi-objective RL problems}

\begin{document}

\twocolumn[
\icmltitle{Benchmarking MOEAs for solving continuous multi-objective RL problems}
           
\icmlsetsymbol{equal}{*}

\begin{icmlauthorlist}
\icmlauthor{Carlos Hernández}{unam}
\icmlauthor{Roberto Santana}{upv}
\end{icmlauthorlist}

\icmlaffiliation{unam}{Applied Mathematics and Systems Research Institute.  National Autonomous University of Mexico. Mexico City. Mexico}
\icmlaffiliation{upv}{Intelligent Systems Group (ISG), University of the Basque Country (UPV/EHU), San Sebastian, Spain}

\icmlcorrespondingauthor{Carlos Hernández}{carlos.hernandez@iimas.unam.mx}
\icmlcorrespondingauthor{Roberto Santana}{roberto.santana@ehu.eus}

\icmlkeywords{Reinforcement learning, MOEAs, benchmarking}

\vskip 0.3in
]



\printAffiliationsAndNotice{\icmlEqualContribution} 

\begin{abstract}
Multi-objective reinforcement learning (MORL) addresses the challenge of simultaneously optimizing multiple, often conflicting, rewards, moving beyond the single-reward focus of conventional reinforcement learning (RL). This approach is essential for applications where agents must balance trade-offs between diverse goals, such as speed, energy efficiency, or stability, as a series of sequential decisions. 

This paper investigates the applicability and limitations of multi-objective evolutionary algorithms (MOEAs) in solving complex MORL problems. We assess whether these algorithms can effectively address the unique challenges posed by MORL and how MORL instances can serve as benchmarks to evaluate and improve MOEA performance. In particular, we propose a framework to characterize the features influencing MORL instance complexity, select representative MORL problems from the literature, and benchmark a suite of MOEAs alongside single-objective EAs using scalarized MORL formulations. Additionally, we evaluate the utility of existing multi-objective quality indicators in MORL scenarios, such as hypervolume conducting a comparison of the algorithms supported by statistical analysis. Our findings provide insights into the interplay between MORL problem characteristics and algorithmic effectiveness, highlighting opportunities for advancing both MORL research and the design of evolutionary algorithms.
\end{abstract}

\section{Introduction}  \label{sec:INTRO}

Multi-objective reinforcement learning (MORL) \cite{Xu_et_al:2020,Hayes_et_al:2022} addresses the problem of optimizing multiple conflicting objectives simultaneously rather than focusing on a single reward signal. While typical single-objective RL algorithms are extensively applied to find optimal policies for agents, there are many scenarios in which the agent pursues diverse goals, or its behavior can be evaluated taking into account different criteria. When a multi-objective scenario is considered, the agent has to balance trade-offs between different goals, which may include speed, energy efficiency, or stability. 

In our opinion, this is an interesting class of problems to address within the multi-objective context. MORL has properties that the community has been interested in recently, such as large-scale, expensive, dynamic, and noisy. Thus, MORL offers a platform to test the algorithms developed in challenging environments that in some cases mimic those of real-world problems.

There is a panoply of approaches to MORL. In this paper, we focus our analysis on evolutionary approaches. On the one hand, we investigate the question of the limits of EAs for solving challenging MORL and if the promising results of evolutionary algorithms on single-objective reinforcement learning translate to the multi-objective case \citep{Salimans_et_al:2017,Such_et_al:2017,majid2023deep}. On the other hand, we are interested in using MORL problems as benchmarks to evaluate different facets of the EAs' performance. We claim that while typical benchmark function can provide a useful characterization of the MOEAs behavior and serve as a platform to advance the design of these algorithms, more challenging problems such MORL can be instrumental in identifying new aspects that need to be improved and open new avenues for research. In this sense, and compared to previous MOEAs benchmarks, the new elements MORL benchmarks provide are the combination of uncertainty, expensive function evaluations and large scale. The main limitation in using these problems for benchmarking MOEAs is the computational time associated to policy evaluation.

We propose a first characterization of elements or features that influence the difficulty that MORL instances pose for MOEAs, select a number of MORL instances, commonly used in the literature, and conduct a comparison of a set of representative MOEAs on these instances. As part of the comparison, we include single-objective EAs that use scalarized variants of the MORL problems. We also provide an analysis of the suitability of existing MO metrics to evaluate the performance of MOEAs.

The paper is organized as follows: Section~\ref{sec:REL_WORK} provides the necessary background on evolutionary RL, multi-objective Markov decision processes, and MOEAs. It also reviews related work relevant to our contribution. Section~\ref{sec:MOEAS} discusses key considerations for applying MOEAs to MORL, including how to evaluate the MOEAs' performance in this scenario. Additionally, it formalizes an MORL instance and identifies the parameters that influence its complexity. Section~\ref{sec:EXPE} presents the experimental benchmark used to evaluate MOEAs, reports the experimental results, and analyzes key findings. Finally, Section~\ref{sec:CONCLU} concludes the paper and outlines potential directions for future research.

\section{Background and Related work}  \label{sec:REL_WORK}
    
    In the following, we present the required background to understand the sequel as well as some related works.
  \subsection{Background}
    Now, we formally define our problem of interest, the so-called Multi-objective Markov Decision Process (MOMDP). A MOMDP is represented by the tuple $<S, A, T, \gamma, \mu, R>$, where
    \begin{itemize}
        \item $S$ is a set of states.
        \item $A$ is a set of actions.
        \item $T:S \times A \times S \rightarrow \left[ 0, 1 \right]$
        \item $\gamma \in \left[0,1) \right.$.
        \item $\mu: S \rightarrow \left[ 0, 1 \right]$ is a probability distribution over initial states.
        \item $R: S\times A\times S \rightarrow \mathbb{R}^k$ is a vectorial reward function for $k$ objectives.
    \end{itemize}
    The main difference between classical MDPs and MOMDPs is that $R$ is a vector function for MOMDPs. In MOMDPs, the agent acts according to a policy $\pi\in\Pi$, where $\Pi$ is the set of all possible policies. A policy $\pi$ is defined as $\pi: S\times A \rightarrow \left[ 0,1 \right]$. Intuitively, one would like to find policies that obtain as much reward as possible. In the following, we will formally define this notion.

    First, in order to evaluate a policy, one can use the value function, which is defined as follows:
    \begin{equation}
        V^\pi = \mathbb{E} \left[ \sum_{i=0}^\infty \gamma^kr_{i+1}|\pi, \mu \right],
        \label{eq:valuefunction}
    \end{equation}
where $r_{i+1}=R(s_i,a_i,s_{i+1})$ is the reward received at step $i+1$. Note that $V^\pi$ is a vector function.

    Analogously to multi-objective optimization, to compare two solutions (policies) $\pi_1 \in \Pi$ and $\pi_2 \in \Pi$, we can use the concept of dominance. A policy $\pi_1$ dominates a policy $\pi_2$ ($\pi_1 \preceq \pi_2$) if $V^{\pi_1} \geq V^{\pi_2}$ and $V^{\pi_1} \neq V^{\pi_2}$. Now, we can define 
    an efficient solution as follows: a policy $\pi^* \in \Pi$ is efficient w.r.t. MOMDP if there is no other policy $\pi\in \Pi$ such that $\pi \preceq \pi^*$.
    
    As one can expect, the solution for such problems is no longer a single policy but rather a set, the so-called Pareto set ($\mathcal{X}$) and its image, the Pareto front ($\mathcal{Y}$).

    \begin{equation}
        \mathcal{X} := \lbrace \pi^* \in \Pi | \not\exists \pi \in \Pi : V^\pi \preceq V^{\pi^*}  \rbrace.
    \end{equation}

    \begin{equation}
        \mathcal{Y} := \lbrace V^{\pi^*} | \pi^* \in \mathcal{X}  \rbrace.
    \end{equation}

    Note that the MOMDP is closely related to multi-objective optimization under uncertainty in the objective functions \cite{beyer2007robust} since we are trying to find solutions for stochastic objective functions.

   \subsection{Related work}
   
   In this section, we present some representative works on evolutionary reinforcement learning (ERL), multi-objective reinforcement learning, and benchmarking of algorithms.
   
    \subsubsection{Evolutionary reinforcement learning}
      The idea of using evolutionary algorithms to solve reinforcement learning problems dates back to the 90s. In \cite{moriarty1999evolutionary}, the authors explored how to adapt evolutionary algorithms to solve this kind of problem. The ideas ranged from a tabular policy representation to using neural networks with good overall results for simple problems. 

      Later, one of the most influential works in ERL investigated the suitability of using evolution strategies (ES)  \cite{Beyer_and_Schwefel:2002} as an alternative to typical RL techniques such as Q-learning and Policy Gradient. In \cite{Salimans_et_al:2017}, the authors evaluated these algorithms on the MuJoCo \cite{Todorov_et_al:2012} and Atari \cite{Mnih_et_al:2015} environment and concluded that ES exhibited a number of advantages as a black box optimization technique. In particular, they identified three advantages: invariance to action frequency and delayed rewards, tolerance to extremely long horizons, and the fact that ES does not need temporal discounting or value function approximation. 
      
      Further, in \cite{Such_et_al:2017}, the authors propose a genetic algorithm (GA) to solve RL problems. The algorithm encodes the neural networks with the random seed and uses them within the evolutionary process of the algorithm. 

      More recently, in \cite{zhu2023survey,majid2023deep}, the authors provide a survey of evolutionary strategies and deep RL. In all cases, the authors have found that EAs can obtain competitive results. Thus, an interesting question is whether EAs' success translates to MORL.    
    
\subsubsection{Multi-objective Markov decision process and MORL approaches}

  MOMDPs \cite{Roijers_et_al:2013} extend the traditional Markov decision process MDP framework to account for multiple, often conflicting, objectives. These models have been widely investigated in recent years due to their applicability in diverse fields, such as robotics, resource management, and healthcare.

MORL has emerged as an important extension of RL to address MOMDP scenarios with multiple conflicting objectives. Unlike single-objective RL, where agents optimize a single reward function, MORL requires balancing trade-offs between competing goals. Researchers have proposed various approaches to handle these challenges, including scalarization methods, Pareto-based strategies, and evolutionary techniques.

Scalarization methods transform multi-objective problems into single-objective ones by weighting and combining objectives into a scalar reward function \cite{Agarwal_et_al:2022,VanMoffaert_et_al:2013}. While this approach simplifies the problem, its effectiveness heavily depends on the choice of weights, which may not always represent user preferences adequately. Scalarization also suffers from a lack of diversity in solutions, as a single policy often dominates others \cite{Roijers_et_al:2013}. Recent advancements attempt to address these limitations by dynamically adjusting weights \cite{Abels_et_al:2019} or using ensemble learning to explore diverse policies.

Pareto-based methods have gained prominence as they aim to approximate the Pareto front, a set of optimal trade-off solutions where improving one objective inevitably worsens another \cite{VanMoffaert_and_Nowe:2014}. These methods often employ techniques like preference elicitation and lexicographic ordering to identify policies that align with user priorities. However, Pareto-based approaches face scalability issues, particularly in high-dimensional objective spaces, where maintaining and comparing Pareto-optimal solutions can be computationally expensive \cite{Hayes_et_al:2022}.

Recent advancements in MORL include the extensive application of multiple deep learning models \cite{Nguyen_et_al:2020} that play different roles in the search for optimal policies.   Another major challenge in MOMDPs is the existence of stochastic components in the evaluation of the reward functions. A number of works have started to address this type of problems \cite{Wang_et_al:2022}

    \subsubsection{Benchmarking of multi-objective evolutionary algorithms}

       Since the early development of MOEAs, the importance of designing effective benchmarks to evaluate algorithmic strengths and weaknesses has been widely recognized. One of the foundational ideas in benchmark design involved identifying problem characteristics that significantly influence the performance of MOEAs \cite{Deb:1999}. Key features initially highlighted included the convexity or non-convexity of the PF, discreteness, and non-uniformity of the PF \cite{zitzler2000comparison}. Deb and colleagues proposed a suite of test problems that were straightforward to construct, scalable in terms of both the number of decision variables and objectives, and for which the exact shape and location of the PF were analytically known \cite{Deb_et_al:2005}. Subsequently, Huband et al. \cite{Huband_et_al:2006} introduced the WFG toolkit, a framework for designing test problems based on a systematic analysis of multi-objective optimization problems (MOP) features. These features include parameter dependencies, modality of the objective functions, Pareto-optimal geometry, and others, which directly characterize the difficulties encountered by optimization algorithms. Recent approaches \cite{Liefooghe_et_al:2023} propose the application of machine learning techniques to discern how the behavior of MOEAs are affected by the MOP features. 
\cite{Liefooghe_et_al:2023}

       Benchmarking continues to be an area of growing interest from several perspectives. For instance, in \cite{bartz2020benchmarking}, the authors analyze the open problems on benchmarking and give several guidelines. Other contributions suggest using real-world problems for benchmarking \cite{kumar2021benchmark} or methodologies to effectively compare algorithms, such as using external archives to store non-dominated solutions \cite{tanabe2017benchmarking}.   Another prominent research direction has been the adaptation of standard benchmarking techniques from single-objective optimization, a field that has achieved a high level of maturity after decades of refinement \cite{Hansen_et_al:2010}, to the multi-objective domain \cite{Brockhoff_et_al:2015}. 
       
       Other research approaches have linked the design of MOP benchmarks with the study of fitness landscape properties \cite{Aguirre_and_Tanaka:2004,Aguirre_and_Tanaka:2007},  and the explicit modeling of interdependencies between decision variables and objectives \cite{Zangari_et_al:2016}. Additionally, research efforts have shifted toward exploring aspects of benchmarking that extend beyond the mere selection or construction of objective functions and their parameters. For example, some authors emphasize metrics and ranking methods for comparing algorithms within a benchmarking context, advocating for the adoption of anytime performance analysis. This perspective facilitates the examination of the relationship between the evaluation budget and the relative performance rankings of EAs \cite{Hansen_et_al:2022,Vermetten_et_al:2024}.

Our work is also closely aligned with studies investigating the characteristics of the search space in ML optimization problems \cite{DuPreez_and_Gallagher:2020,Pimenta_et_al:2020,Schneider_et_al:2022,Liefooghe_et_al:2024}, particularly those focusing on the landscape of neural network weights within ML tasks \cite{Bosman_et_al:2023,Moses_et_al:2021,Neri_and_Turner:2023}. Such problems exhibit unique characteristics that distinguish them from traditional MOEA benchmarks and merit further investigation. Notably, these problems often involve rapid scalability in the number of decision variables, making them computationally challenging. Additionally, solutions tend to be computationally expensive due to the inference requirements on large datasets. Moreover, the quality of solutions can be evaluated based on multiple criteria, underscoring their inherently multi-objective nature.

\section{MORL on the multi-objective optimization framework} \label{sec:MOEAS}  

In this section, we explore a methodology to adapt MORL problems to MOPs and the pertinence of known quality indicators for this context. We also discuss the question of characterizing MORL instances.

\subsection{Solving MOMDPs with MOEAs}  
As is common when solving a problem with evolutionary algorithms, there are two main challenges to address: choosing a suitable representation of the problem and how to assess the performance of an individual.

For the first challenge, we follow the work presented in \cite{moriarty1999evolutionary}. As mentioned before, in this work, the authors proposed to use neural networks to represent a policy. Thus, we have $\pi_\theta: S\times A \rightarrow \left[ 0, 1 \right]$ where $\theta \in \mathbb{R}^n$ and $n$ is the number of neural network parameters. In this work, we restrict ourselves to fully connected feedforward-type networks. It is important to note that although other algorithms exist to optimize neural network parameters, such as stochastic gradient descent, in single-objective reinforcement learning, evolutionary algorithms have shown competitive performance. Thus, it is interesting to explore whether it is still valid for multi-objective reinforcement learning.

For the second challenge, note that we can use the Equation \ref{eq:valuefunction} as our objective function. However, this equation requires to compute the expectation of the discounted sum of rewards. From the evolutionary computation perspective, it is a function subject to uncertainty. There are several methods to deal with such problems. In our case, we decided to use one of the simplest methods, Monte Carlo sampling. That is, we compute several samples for the objective function for each solution in the population and compute the empirical mean, similar to \cite{deb2006introducing}. In our opinion, it provides a good starting point for our comparison.

Although the adaptation can be seen as straightforward, these kinds of problem can be high-dimensional from using neural networks; they have potentially expensive function evaluations since we are required to complete an episode from the MOMDP, and the problem is subject to uncertainty in the objective function. In our opinion, these kinds of difficulty make MORL problems interesting to study and benchmark MOEAs on.

\subsection{Quality indicators of MORL}

Another essential aspect of MORL is the evaluation of algorithmic performance. Although many performance indicators exist for multi-objective optimization \cite{Zitzler_et_al:2000, coello2007evolutionary}, most of them have been proposed for deterministic problems. Thus, these metrics may not fully capture the practical requirements of MORL, such as adaptability to changes in objectives or robustness under varying conditions. 

For our work, we decided to adapt classical performance indicators, such as Hypervolume, Generational Distance (GD), and Inverted Generational Distance (IGD), by applying them to the approximation of the expected Pareto front found by algorithms.

We acknowledge that recent studies have proposed alternative evaluation frameworks that consider these aspects, potentially offering a more comprehensive understanding of MORL algorithm effectiveness \cite{Hayes_et_al:2022,branke2024performance}. However, a detailed analysis of the indicators should be performed for use in the MORL context.

\subsection{Characterizing MORL instances}

In the context of MOP benchmarking, a fundamental challenge in designing benchmark instances is identifying the key features that characterize the problem. In MORL, as in other optimization problems in ML, it is difficult to determine a priori which characteristics, such as multimodality or PF geometry, are most relevant. Since reward functions emerge from interactions with the environment, and the mapping between decision variables is mediated by the inference process, predefining specific properties of the objective space or fitness landscape is inherently challenging.

We propose distinguishing between two types of features that can aid in characterizing MORL instances:  

   \begin{enumerate}
     \item Intrinsic problem features: They define the fundamental nature of the problem, such as the number of reward functions to optimize, the degree of conflict of objectives, whether the environment provides partial or full observability,  the level of randomness in state transitions, and the complexity of the environment.  
     
      \item Learning features: These refer to characteristics influenced by the learning process, including the number of model parameters (weights and biases), the loss function used for training, and the number of training episodes.  
      
   \end{enumerate}

Intrinsic problem features can be further categorized based on whether they stem from the agent's characteristics, the environment's specificity, or their interaction dynamics. Similarly, learning features can be subdivided into those defining the neural network architecture and those related to objective function computation. A more granular identification of these features, combined with systematic variation, allows for a comprehensive assessment of MOEAs across diverse scenarios. While these features differ from conventional MOEA benchmarking criteria, they can help in designing more realistic and diverse problem instances, also contributing to uncovering unexpected patterns in evolutionary algorithm behavior, as evidenced by our experimental results.

Many of the existing benchmarks focus on a few difficulties. For instance, multi-modality, expensive, large-scale, many objective, among others. This allows us to isolate the properties of interest and control them. In the other hand, the MORL benchmark offers problems that combine several of these difficulties in the same problem. In addition, they can represent, in many cases, scenarios closer to those of real-world problems.

\section{Experiments}  \label{sec:EXPE}
  
  The goal of our experiments is to evaluate how well MOEAs can solve continuous MORL problems. We also examine how different aspects of MORL instances affect the performance of these algorithms. Specifically, our aim is to answer these questions:
  
  \begin{itemize}
      \item  Do MOEAs outperform single-objective EAs that use a scalarized version of the problem?
      \item  How does the length of the agent's learning period (number of episodes) affect MOEA performance?
      \item Does increasing the complexity of the agents (by adding more parameters to the neural network) influence MOEA performance?

  \end{itemize}
  
  Our analysis does not focus on the particular computational time spent by the different MOEAs. We assume that the most expensive part of the algorithms is the function evaluation and assign the same budget of evaluations to all algorithms for a fair comparison. Efficiency is therefore evaluated in terms of the quality of the PF approximations obtained with this number of evaluations. 
  
 \subsection{Experimental framework}

   The design of the experimental framework takes into account the research questions previously introduced. The main elements of the configurations are as follows: 

   \begin{itemize}
      \item Population size: 50
      \item Number of generations = $\{25,100\}$
      \item Single-Objective EAs: DE, PSO, GA
      \item MOEAs: NSGA2, SMSEMOA, NSGA3, RNSGA2, SPEA2
      \item Neural network architecture of the agent: Three hidden layers, $n\_layer1 = 4$, $n\_layer2 = \{4,10\}$,  $n\_layer3 = 4$.
      \item Number of episodes:  $\{1,5,10\}$      
  \end{itemize}

    As a baseline, we selected three single-objective EAs and defined the single-objective function by scalarizing all objectives. Scalarization is implemented by assigning each scalar a weight equal to the inverse of the number of objectives.  By evaluating both single-objective EAs and MOEAs, we aim to gain a more comprehensive understanding of how different search strategies affect the quality PF approximations in MORL. The selected EAs—both single- and multi-objective—were chosen to represent a diverse set of representative algorithmic approaches.
    
    By testing  single-objective EAs and MOEAs we can have a more complete idea of how different search strategies influence the quality of PF approximations for MORL.  The particular choice of single-objective EAs and MOEAs have been done aiming at a variety of representative EAs. 
    
    The main configuration for the algorithm in our experiments includes a population of $50$ individuals over $25$ generations and $5$ episodes. In addition, we tested the algorithm with a varying number of episodes and evaluated its performance with longer generations.     
  
      To evaluate the quality of a PF approximations produced by the MOEAs, we first computed a reference PF approximation from the results of all algorithms for a given problem. The reference PF is used to estimate the performance of each individual algorithm using a given metric and taking the reference PF as ideal. PFs for individual algorithms were computed from the union of all executions of the given algorithm. 

      The code used for the experiments as well as the results obtained are available from \url{https://github.com/rsantana-isg/MORL_benchmark}.
  
\subsubsection{Test problems}

 We use the MuJoCo (Multi-Joint Dynamics with Contact) physics engine \cite{Todorov_et_al:2012} to simulate the agents and evaluate the quality of the solutions. In this physics environment, the tasks simulate complex robotic movements, where agents must learn to navigate and interact with their environment while considering multiple criteria for success.

 We focus on the following problems: \emph{mo-hopper-v4}, \emph{mo-halfcheetah-v4}, \emph{mo-walker2d-v4}, \emph{mo-ant-v4}, and \emph{mo-humanoid-v4}. The first and fourth problems have three reward functions, while the other problems are bi-objective. A more detailed description of these problems is presented in Section~1 of the Supplementary Material.

\subsection{Performance in terms of HV}

\begin{figure}[htbp]
    \centering
    \includegraphics[width=7.5cm]{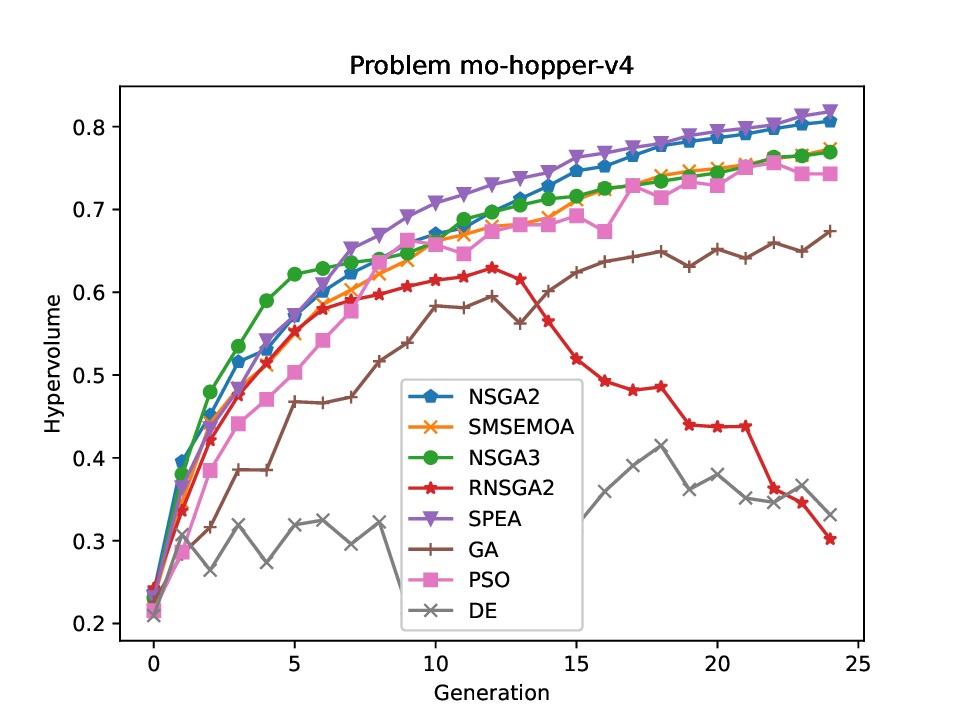}
    \caption{Evolution of the HV for the problem \texttt{mo-hopper-v4} with $n\_episodes=5$.}      
    \label{fig:RESULTS_CM_0_NEP_5}
\end{figure}

\begin{figure}[htbp]
    \centering
    \includegraphics[width=7.5cm]{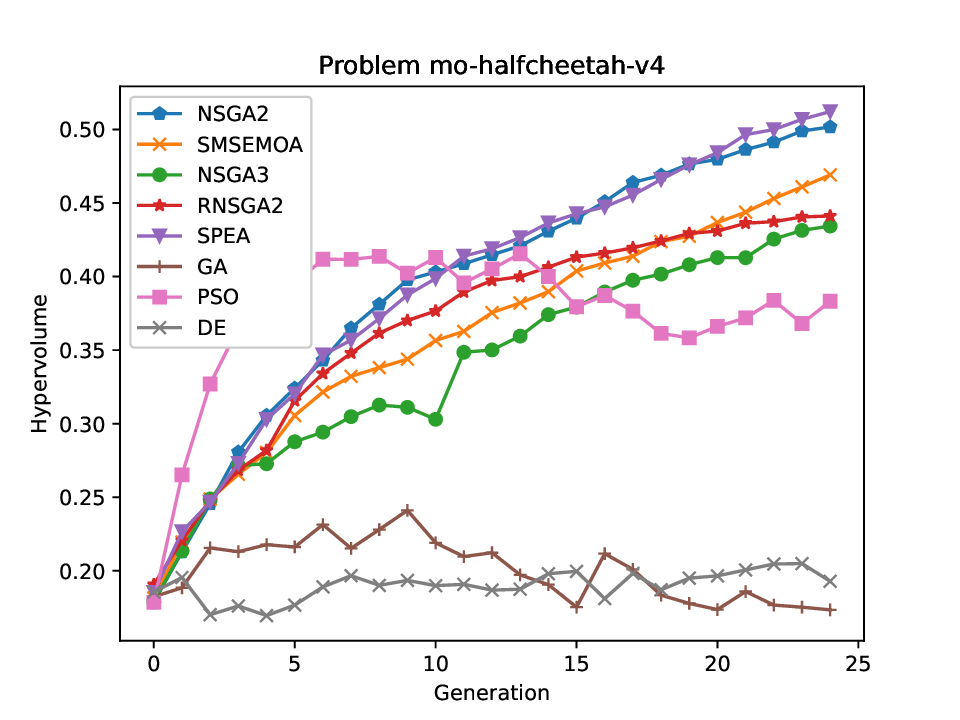}
    \caption{Evolution of the HV for the problem \texttt{mo-halfcheetah-v4} with $n\_episodes=5$.}       
    \label{fig:RESULTS_CM_1_NEP_5}
\end{figure}

Figures~\ref{fig:RESULTS_CM_0_NEP_5} and~\ref{fig:RESULTS_CM_1_NEP_5} illustrate the performance of all EAs for the problems \texttt{mo-hopper-v4} and \texttt{mo-halfcheetah-v4}, respectively. These results were obtained using the baseline configuration: $n\_episodes=5$, $n\_layer2=4$, and a maximum of $ngen=25$ generations. Each figure presents the evolution of the HV as the number of generations increases. Results for the remaining benchmark problems are provided in Section~2 of the Supplementary Material.

An analysis of the results for \texttt{mo-hopper-v4} demonstrates that SPEA2 and NSGA2 behave similarly for this 3-objective optimization problem. They exhibit steady improvements in performance as the number of evaluations increases. Among SOEAs, PSO demonstrates performance comparable to SMS-EMOA and NSGA3, whereas GA, and particularly DE, improve much more slowly. Interestingly, RNSGA2 displays a unique behavior: it improves HV up to generation $12$ but then gradually deteriorates, reducing the quality of the PF approximations over time.

The results for the \texttt{mo-halfcheetah-v4} problem, shown in Figure~\ref{fig:RESULTS_CM_1_NEP_5}, confirm a significant performance gap between MOEAs and SOEAs. DE and GA fail to show any substantial improvement in HV as the number of evaluations increases. While PSO exhibits a strong initial performance, outperforming both SOEAs and MOEAs, it stagnates after five generations and is eventually outperformed by all MOEAs. The best-performing algorithms for this problem are SPEA2 and NSGA2, which consistently deliver superior results.

The results for the remaining three benchmark problems, presented in Section~2 of the Supplementary Material, reveal additional insights:
\begin{itemize}
    \item For 3-objective MORL problems,NSGA2 and SPEA2 produce the best results, although SMSEMOA has a competitive performance.
    \item For 2-objective MORL problems, SPEA2 and NSGA2 achieve the best performance on \texttt{mo-walker2d-v4}.
    \item For \texttt{mo-humanoid-v4}, RNSGA2, PSO, and GA emerge as the top-performing algorithms, highlighting that SOEAs can occasionally rival or surpass MOEAs under specific conditions.
\end{itemize}

 \subsection{PFs found for the MORL problems}

From the results of all EAs for $n\_episodes=5$ and $ngen=25$ generations, we have computed the PF approximations for all benchmark problems. These PF approximations provide valuable insights into the characteristics of the optimization problems and serve as reference sets for evaluating additional metrics of MOEA performance. 

For the sake of clarity and interpretability, we focus our analysis on the 2-objective \texttt{mo-halfcheetah-v4} and \texttt{mo-humanoid-v4}, as the visualization of PFs in two dimensions facilitates a more intuitive understanding. Furthermore, these two problems exhibit distinct differences in the behavior of EAs, making them ideal for comparative analysis. The PFs for the remaining three benchmark problems are provided in Section~3 of the Supplementary Material.

\begin{figure}[htbp]
    \centering
    \includegraphics[width=7.5cm]{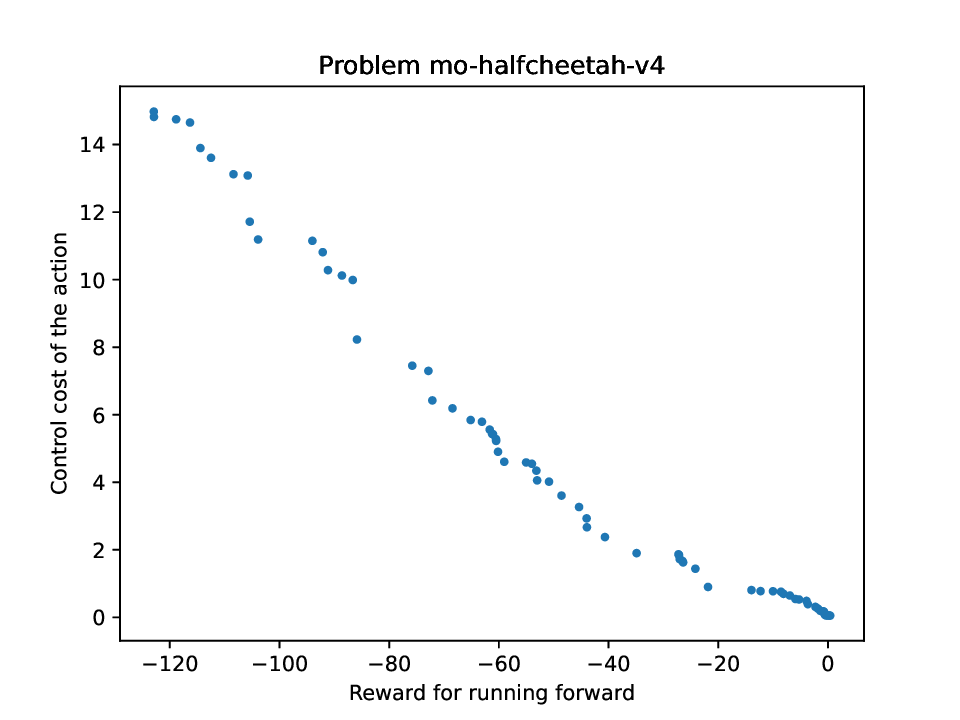}
    \caption{Pareto front for the problem \texttt{mo-halfcheetah-v4} with $n\_episodes=5$.}       
    \label{fig:PF_CM_1_NEP_5}
\end{figure}  

\begin{figure}[htbp]
    \centering
    \includegraphics[width=7.5cm]{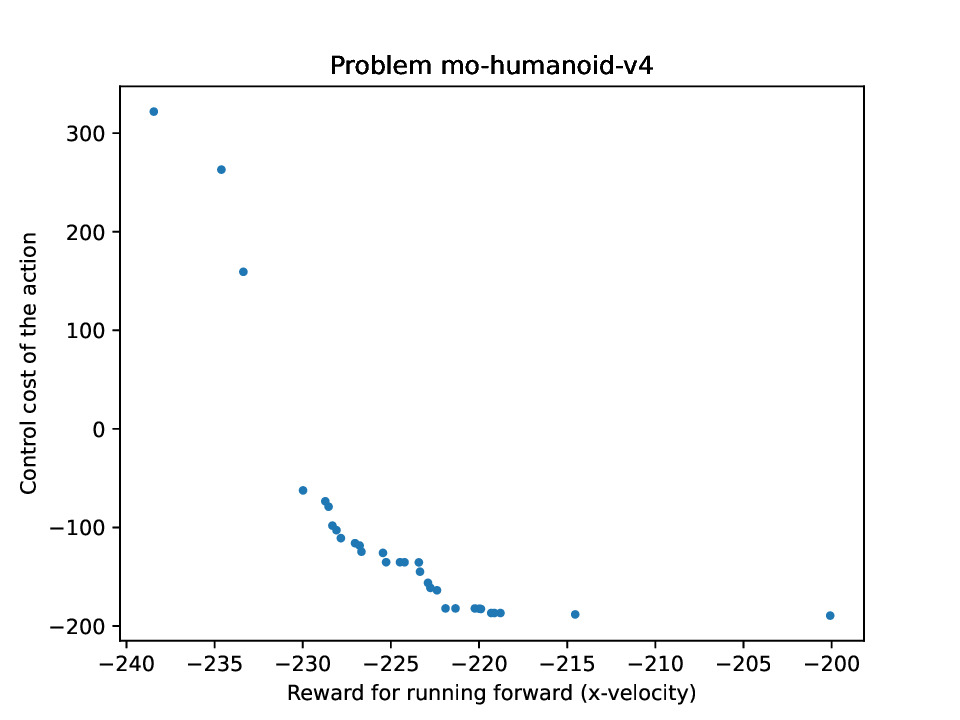}
    \caption{Pareto front for the problem \texttt{mo-humanoid-v4} with $n\_episodes=5$.}       
    \label{fig:PF_CM_4_NEP_5}
\end{figure}

Figures~\ref{fig:PF_CM_1_NEP_5} and~\ref{fig:PF_CM_4_NEP_5} depict the PF approximations for the problems \texttt{mo-halfcheetah-v4} and \texttt{mo-humanoid-v4}, respectively. The figures are the non-dominated solutions of the union of all executions with all algorithms. The PF for \texttt{mo-halfcheetah-v4} exhibits a well-distributed set of solutions along the front, suggesting that the MOEAs are capable of finding diverse solutions spanning the trade-off surface.

In contrast, the PF approximation found by the MOEAs for \texttt{mo-humanoid-v4} is concentrated in a narrower region of the objective space, indicating significant differences in problem characteristics. This concentration could suggest the PF is highly convex, which could explain why most methods focus on the center of the front. Note that there is a need to analyze the fronts further and attempt other techniques, such as local search and landscape analysis, to validate this hypothesis.

\subsection{Analysis of other PF metrics}

\begin{figure}[htbp]
    \centering
    \includegraphics[width=7.5cm]{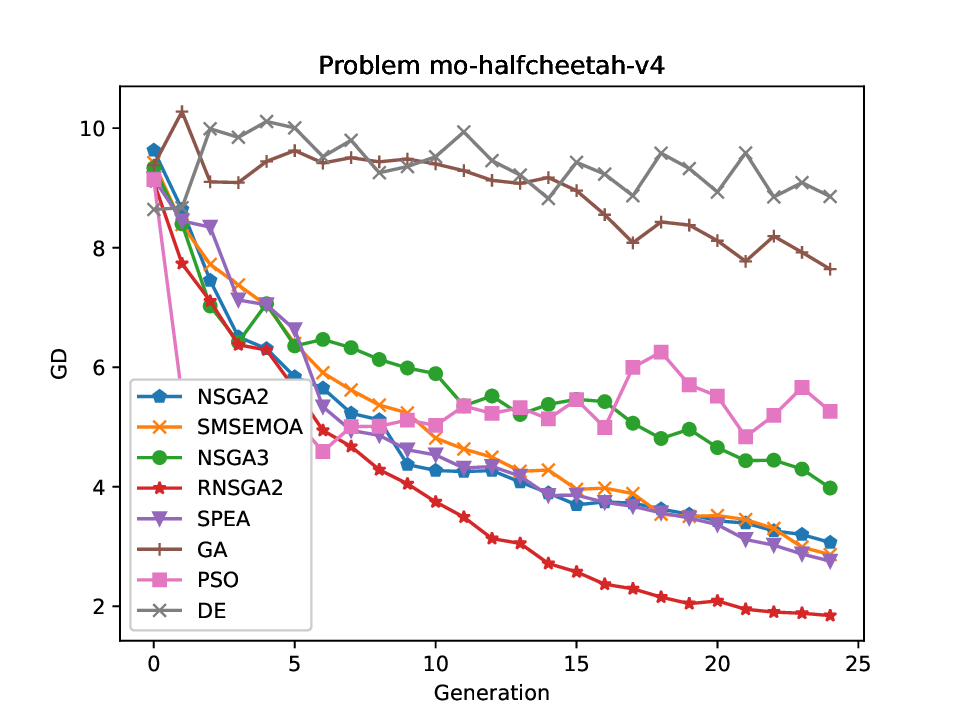}
    \caption{Evolution of the GD metric for the problem \texttt{mo-halfcheetah-v4} with $n\_episodes=5$.}       
    \label{fig:GD_CM_1_NEP_5}
\end{figure}

\begin{figure}[htbp]
    \centering
    \includegraphics[width=7.5cm]{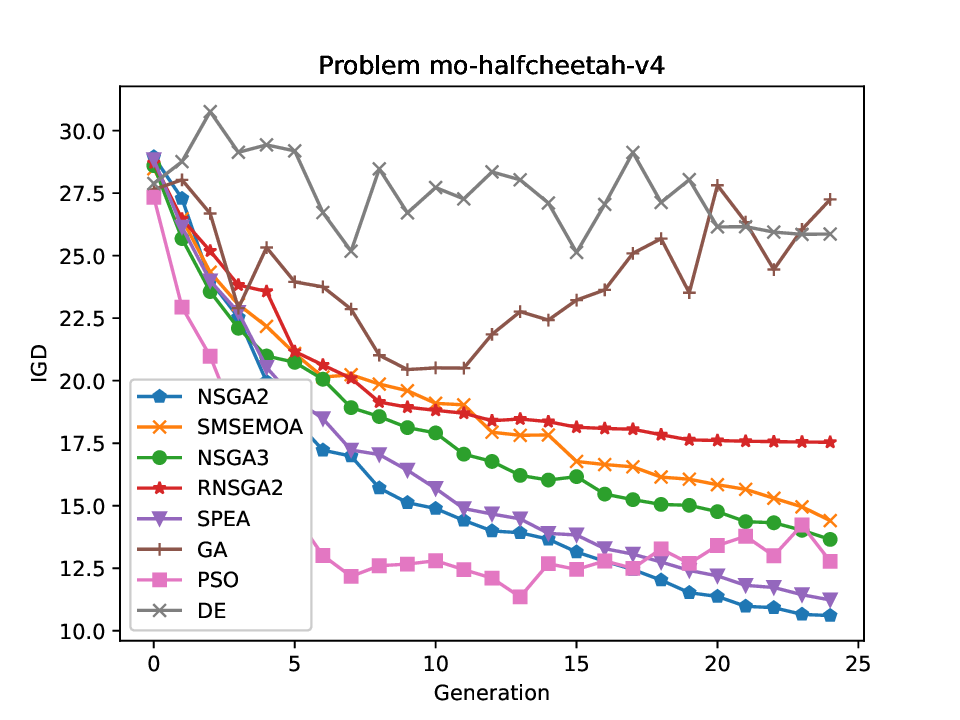}
    \caption{Evolution of the IGD metric for the problem \texttt{mo-halfcheetah-v4} with $n\_episodes=5$.}       
    \label{fig:IGD_CM_1_NEP_5}
\end{figure}

In addition to HV, we evaluated the performance of the algorithms using two additional metrics: Generational Distance (GD) and Inverted Generational Distance (IGD). Figures~\ref{fig:GD_CM_1_NEP_5} and~\ref{fig:IGD_CM_1_NEP_5} show the GD and IGD metrics, respectively, for the \texttt{mo-halfcheetah-v4} problem. Detailed results for other problems and metrics are provided in Section~4 of the Supplementary Material.

The GD metric indicates that RNSGA2 is the best-performing algorithm for this problem. However, when evaluated using the IGD metric, SPEA2 and NSGA2 emerge as the most effective algorithms. This discrepancy highlights the multifaceted nature of evaluating MOEAs and emphasizes the importance of using diverse performance metrics to capture different aspects of algorithmic behavior. By employing multiple metrics on the MORL benchmark, we gain a broader perspective on the performance of MOEAs, enabling a more comprehensive characterization of their strengths and limitations.

A notable observation from the IGD analysis for other problems (figures in Section~4 of the Supplementary Material) is the high efficiency of the PSO algorithm according to this metric. This suggests that PSO can effectively learn a good approximation of the true PF in some scenarios.

\subsection{Statistical analysis of the differences}

We performed a comprehensive statistical analysis to assess the existence of significant differences among the algorithms based on the selected performance metrics. This analysis considered the metrics obtained from the final populations in the $10$ independent runs executed for all algorithms, using the baseline benchmark configuration ($n\_episodes=5$).

\begin{figure*}[htbp]
    \centering
    \includegraphics[width=5.7cm]{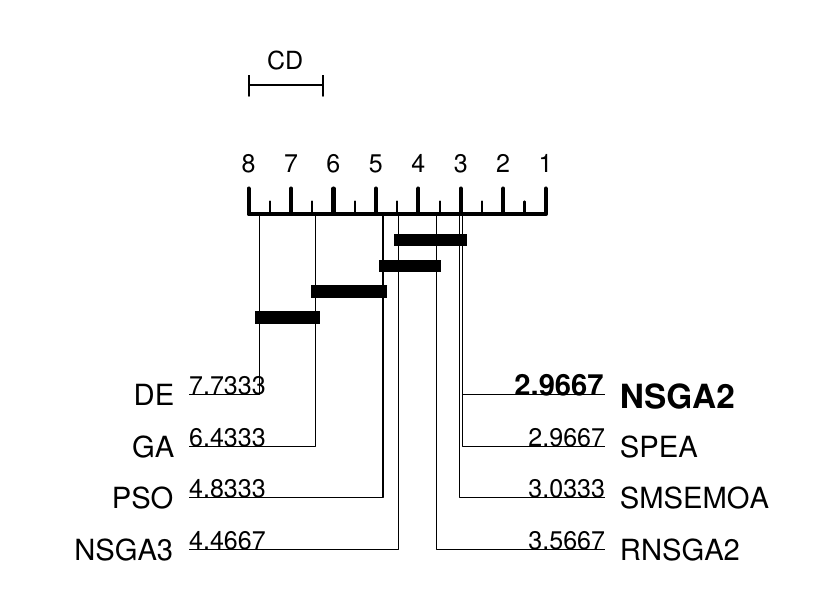}    
    \includegraphics[width=5.7cm]{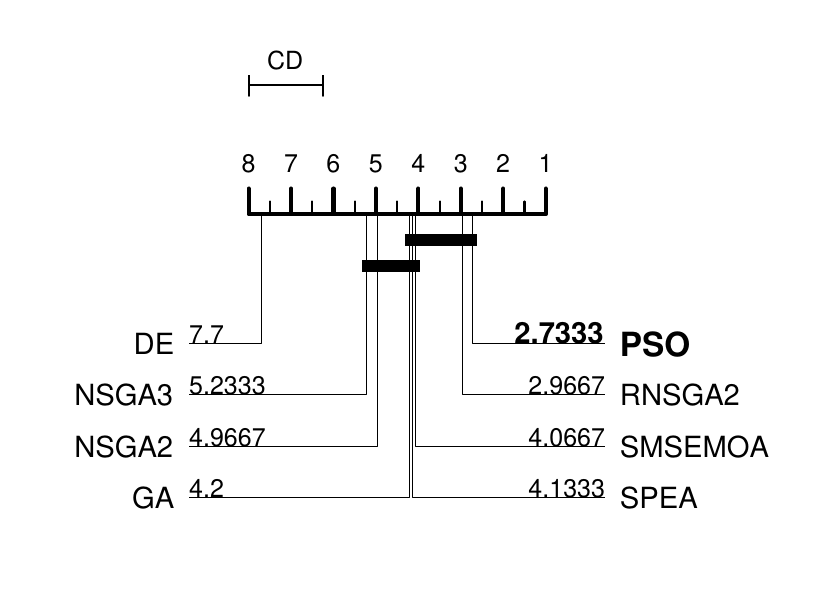}
    \includegraphics[width=5.7cm]{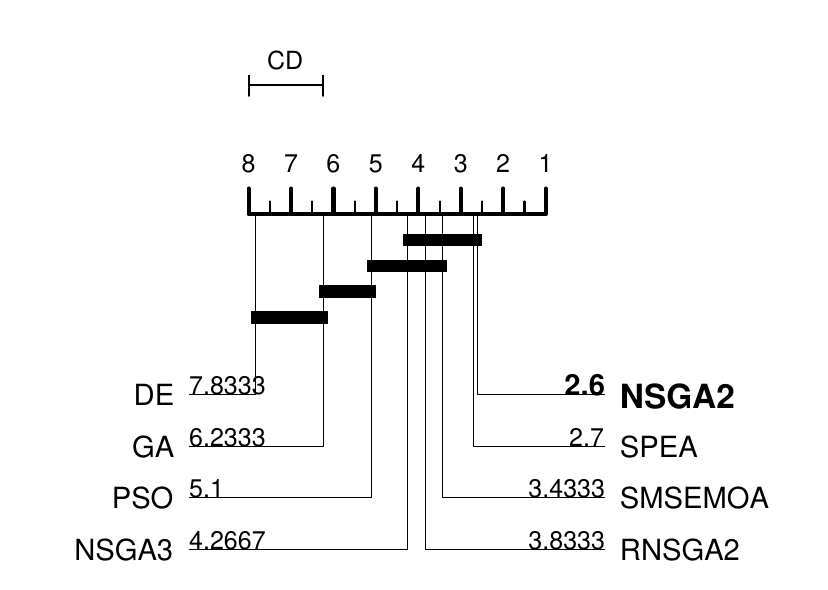}
    \caption{Critical difference diagrams for HV, GD, and IGD metrics computed for problems with 2 objectives.}      
    \label{fig:DIFF_DIAGRAMS_K2}

\end{figure*}

\begin{figure*}[htbp]
    \centering
    \includegraphics[width=5.7cm]{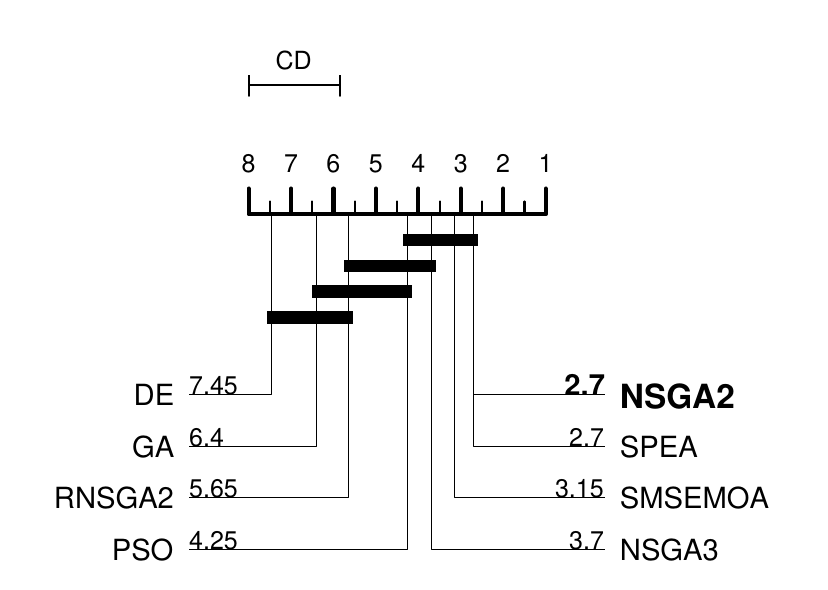}    
    \includegraphics[width=5.7cm]{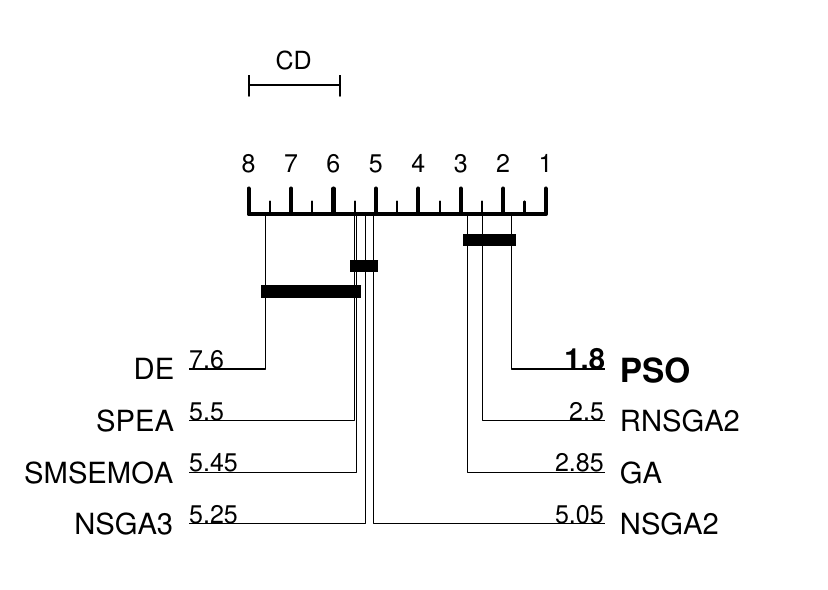}
    \includegraphics[width=5.7cm]{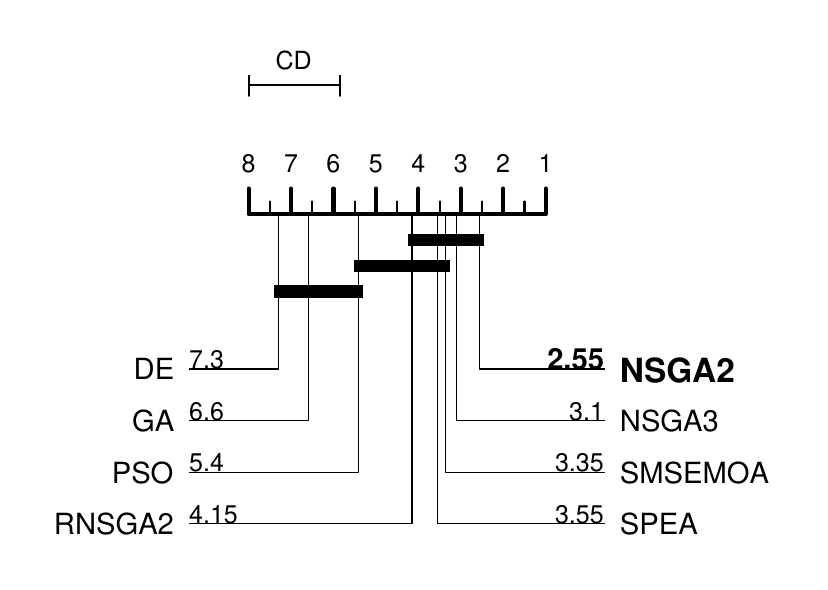}
    \caption{Critical difference diagrams for HV, GD, and IGD metrics computed for problems with 3 objectives.}      
    \label{fig:DIFF_DIAGRAMS_K3}
\end{figure*}

Figures~\ref{fig:DIFF_DIAGRAMS_K2} and~\ref{fig:DIFF_DIAGRAMS_K3} respectively present the critical difference diagrams calculated for problems with two and three objectives. The figures are the result of applying the Friedman test and Nemenyi as post-hoc with an alpha of 0.05. The figures show information about the statistical differences among the algorithms for the HV, GD, and IGD metrics. 
 

 Considering problems with two objectives and in terms of the HV metric, the best-performing algorithms are NSGA2 and SPEA2, although their performance is not significantly different from the performance of the other MOEAs. A similar ranking of MOEAs is achieved if the IGD metric is used.  However, notable differences exist between the group formed by NSGA2 and SPEA2 and that formed by all SOEAs. If the GD metric is considered, then PSO is the best method, although it is not significantly different from the other three MOEAs, including SPEA2. Considering all the metrics, SPEA2 outstands as an algorithm that is not significantly outperformed by any other EA.

 For the three objectives, NSGA2 achieves the best results when considering the HV and IGD metrics. However, it is not significantly different from all other MOEAs except RNSGA2. Also, in 3-objectives problems, PSO has a remarkable performance for the GD metric, with no significant differences between RNSGA2 and GA. This outcome is likely attributable to the tendency of PSO and other SOEAs to effectively explore smaller regions of the PF, as dictated by the scalarized objective functions they optimize.

\subsection{Results for the optimization of the scalarized objective}

\begin{figure}[htbp]
    \centering
    \includegraphics[width=7.5cm]{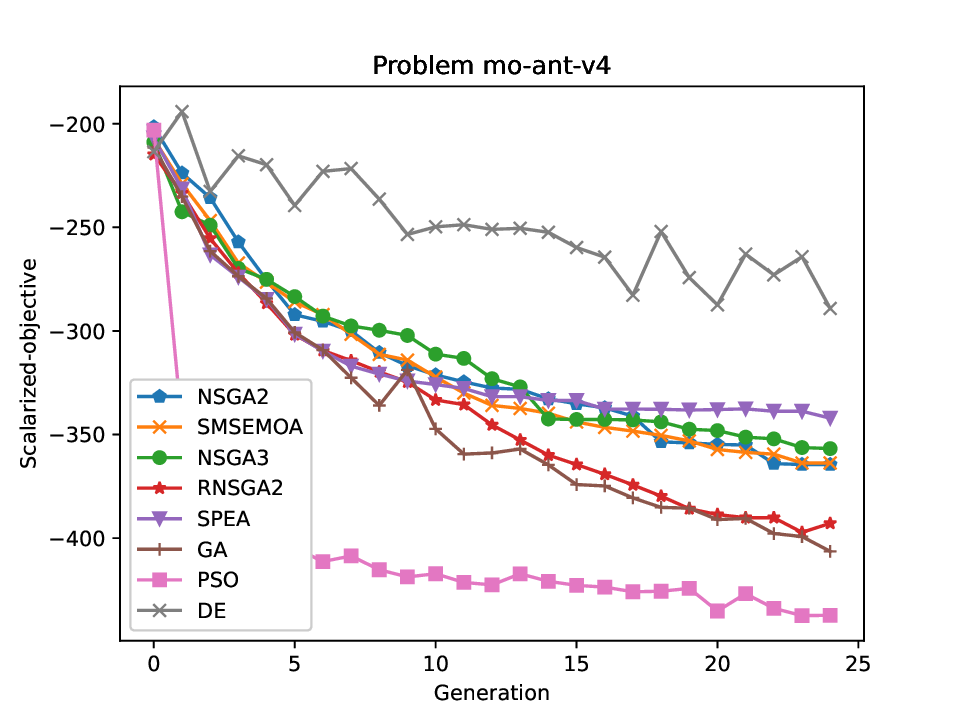}
    \caption{Evolution of the scalarized objective for the problem \texttt{mo-hopper-v4} with $n\_episodes=5$.}    
    \label{fig:SCA_CM_0_NEP_5}
\end{figure}

\begin{figure}[htbp]
    \centering
    \includegraphics[width=7.5cm]{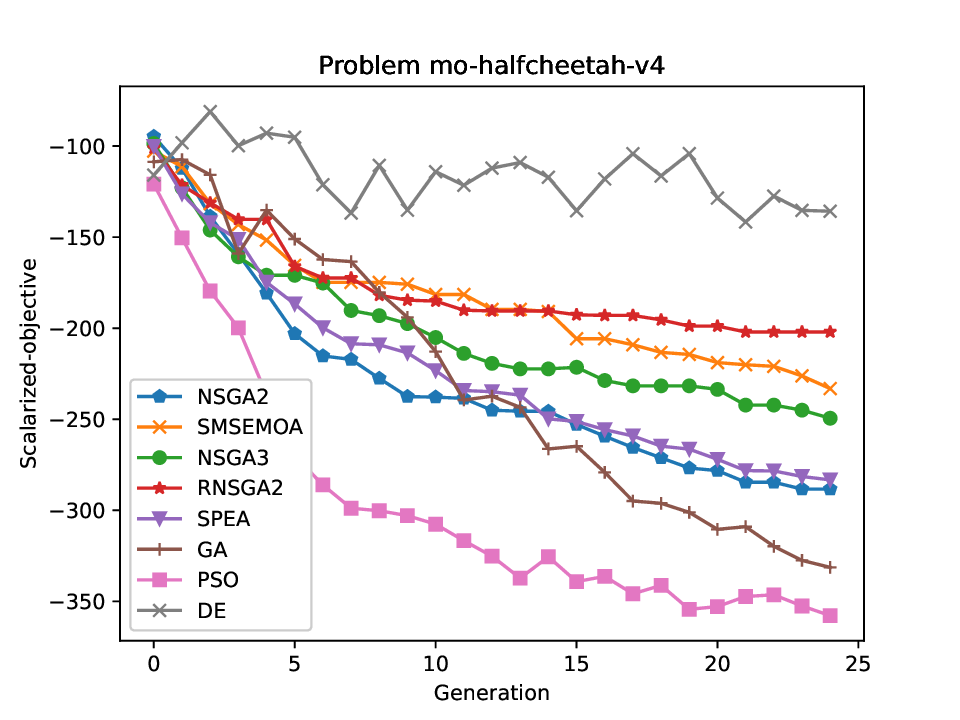}
    \caption{Evolution of the scalarized objective for the problem \texttt{mo-halfcheetah-v4} with $n\_episodes=5$.}       
    \label{fig:SCA_CM_1_NEP_5}
\end{figure}

While metrics such as HV, GD, and IGD are well-suited for evaluating the quality of the PF approximations, they do not align with the specific optimization objectives of SOEAs, which aim to maximize a scalarized reward. To address this, we analyzed the evolution of the scalarized objective function across all EAs. Figures~\ref{fig:SCA_CM_0_NEP_5} and~\ref{fig:SCA_CM_1_NEP_5} illustrate these results for the problems \texttt{mo-hopper-v4} and \texttt{mo-halfcheetah-v4}, respectively. Additional results for other problems are presented in Section~5 of the Supplementary Material.

From figures~\ref{fig:SCA_CM_0_NEP_5} and~\ref{fig:SCA_CM_1_NEP_5}, it is evident that PSO and GA demonstrate rapid improvements in the scalarized objective function during the initial stages of the search. In these two problems, no other algorithm outperforms PSO and GA in terms of the scalarized objective. 

These findings underscore an important distinction that is also valid when benchmarking MOEAs in MORL instances: optimizing scalarized rewards may improve the specific objective values but does not necessarily lead to a well-distributed set of non-dominated solutions. This observation aligns with earlier analyses of PF approximations, where MOEAs demonstrated superior performance in exploring and approximating diverse trade-off solutions.

\subsection{Investigating the Behavior of MOEAs for Other MORL Benchmark Configurations}

\begin{table*}[htbp]
   \tiny
    \centering   
    \begin{tabular}{@{}c|rrrrr|rrrrr@{}}
        \toprule
      Configuration    &  \multicolumn{5}{c|}{HV  mean(std)}  &  \multicolumn{5}{c}{IGD mean(std)} \\ \hline 
         ($e$, $l$, $g$) & PSO & NSGA2 & SMSEMOA  & RNSGA2 & SPEA2    & PSO & NSGA2 & SMSEMOA & RNSGA2 & SPEA2    \\ 
        \midrule        
        (1,4,25)    & 0.452 (0.026) & 0.582 (0.039) & 0.588 (0.037) & 0.383 (0.086) &  {\bf{0.605}} (0.062) & 0.936 (0.100) & 0.481 (0.059) &  {\bf{0.456}} (0.065) & 1.034 (0.319) & 0.477 (0.084) \\
    (5,4,25)    & 0.743 (0.119) & 0.807 (0.066) & 0.773 (0.071) & 0.302 (0.217) &  {\bf{0.818}} (0.062) & 14.593 (6.155) & 4.786 (0.797) & 5.357 (0.711) & 15.811 (6.294) &  {\bf{4.645}} (0.632) \\
    (10,4,25)   & 0.802 (0.080) & 0.864 (0.037) & 0.857 (0.052) & 0.495 (0.210) &  {\bf{0.876}} (0.090) & 16.466 (5.050) & 5.586 (0.910) & 5.775 (0.638) & 13.578 (5.634) &  {\bf{5.573}} (0.872) \\
    (10,10,25)  & 0.705 (0.093) &  {\bf{0.724}} (0.050) & 0.719 (0.050) & 0.440 (0.199) & 0.717 (0.038) & 15.459 (5.776) & 9.404 (1.845) & 8.922 (0.703) & 16.502 (7.258) &  {\bf{8.344}} (1.210) \\
    (10,10,100) & 0.744 (0.038) & 0.880 (0.041) &  {\bf{0.907}} (0.044) & 0.137 (0.078) & 0.842 (0.046) & 18.702 (4.732) & 5.711 (1.085) &  {\bf{4.665}} (0.891) & 32.410 (7.460) & 5.129 (1.105) \\
    (30,10,25)  & 0.878 (0.076) & 0.856 (0.020) & 0.831 (0.060) & 0.662 (0.276) &  {\bf{0.889}} (0.077) & 14.617 (4.381) & 9.564 (1.082) & 9.503 (2.076) & 14.784 (9.595) &  {\bf{8.550}} (1.234) \\
    \end{tabular}
     \caption{Performance comparison for problem \texttt{mo-hopper-v4} using different configurations of the benchmark parameters, where $e$,$l$, $g$ correspond to $n\_ep$, $n\_layer2$, $n\_gen$, respectively.}
    \label{tab:CONF_COMP_PROBLEMS_0}
\end{table*}

\begin{table*}[htbp]
   \tiny
    \centering   
    \begin{tabular}{@{}c|rrrrr|rrrrr@{}}
        \toprule
      Configuration    &  \multicolumn{5}{c|}{HV  mean(std)}  &  \multicolumn{5}{c}{IGD  mean(std)} \\ \hline 
         ($e$, $l$, $g$) & PSO & NSGA2 & SMSEMOA  & RNSGA2 & SPEA2    & PSO & NSGA2 & SMSEMOA & RNSGA2 & SPEA2    \\ 
        \midrule        
          (1,4,25)    & 0.440 (0.236) & {\bf{0.614}} (0.078) & 0.575 (0.077) & 0.617 (0.096) & 0.556 (0.077) & 0.602 (0.307) & 0.552 (0.207) & 0.581 (0.159) & {\bf{0.511}} (0.180) & 0.610 (0.198) \\
    (5,4,25)    & 0.383 (0.171) & {\bf{0.502}} (0.048) & 0.469 (0.031) & 0.441 (0.026) & 0.512 (0.044) & 12.849 (3.483) & {\bf{10.700}} (2.797) & 14.549 (2.336) & 17.663 (3.605) & 11.342 (3.392) \\
    (10,4,25)   & 0.329 (0.162) & 0.533 (0.047) & 0.518 (0.044) & 0.463 (0.027) & {\bf{0.551}} (0.079) & 18.434 (11.184) & 6.318 (1.130) & 7.012 (1.646) & 9.599 (1.760) & {\bf{6.121}} (1.757) \\
    (10,10,25)  & 0.317 (0.129) & {\bf{0.461}} (0.066) & 0.416 (0.054) & 0.400 (0.031) & 0.441 (0.032) & 20.367 (6.507) & {\bf{15.171}} (4.182) & 17.167 (3.430) & 20.722 (1.745) & 16.484 (3.007) \\
    (10,10,100) & 0.269 (0.126) & {\bf{0.667}} (0.036) & 0.617 (0.069) & 0.251 (0.074) & 0.656 (0.053) & 28.488 (9.630) & {\bf{6.681}} (2.198) & 9.168 (3.497) & 36.981 (6.378) & 7.594 (3.513) \\
       \bottomrule
    \end{tabular}
     \caption{Performance comparison for problem \texttt{mo-halfcheetah-v4} using different configurations of the benchmark parameters, where $e$, $l$, $g$ correspond to $n\_ep$, $n\_layer2$, $n\_gen$, respectively.}
    \label{tab:CONF_COMP_PROBLEMS_1}
\end{table*}

The previous results focused on algorithm performance with $n\_episodes = 5$. To better understand the impact of key parameters in MORL instances, we analyzed a subset of EAs under different configurations.

Tables~\ref{tab:CONF_COMP_PROBLEMS_0} and~\ref{tab:CONF_COMP_PROBLEMS_1} compare the algorithms in terms of HV and IGD at the final population for \texttt{mo-hopper-v4} and \texttt{mo-halfcheetah-v4}, respectively. Notably, changes in $n\_episodes$ directly affect the objective function, as it is computed as the average reward over episodes. Similarly, increasing the neural network's parameter count expands the search space, making optimization more challenging. Thus, results are primarily comparable within each row of the table.

Table~\ref{tab:CONF_COMP_PROBLEMS_0} highlights significant ranking shifts due to MORL configuration changes. For instance, SMSEMOA achieves the best HV and IGD in the $(10,10,100)$ setting but ranks below SPEA2 for the  $(30,10,25)$  configuration. Similarly, Table~\ref{tab:CONF_COMP_PROBLEMS_1} shows that while SPEA2 outperforms NSGA2 in HV and IGD for $(10,4,25)$, NSGA2 becomes the top performer when the neural network complexity increases (i.e., more decision variables), particularly when $n\_episodes = 10$ and $n\_layers = 10$, regardless of whether $25$ or $100$ EA generations are used.

Overall, our findings indicate that algorithm performance varies depending on the MORL instance characteristics. SPEA2 has a generally better performance for \texttt{mo-hopper-v4}, and NSGA2 is better for \texttt{mo-halfcheetah-v4}. Some methods are more sensitive to the number of episodes, while others struggle with increased model complexity. Additional results on the evolution of these five EAs across different problems, MORL configurations, and metrics can be found in Section~6 of the Supplementary Material.


 \subsection{Discussion and Lessons Learned}

Benchmarking MOEAs on the MORL problems has led to several important findings:

\begin{itemize}
    \item \textbf{PSO as a promising candidate for MORL tasks}: PSO demonstrates remarkable performance in the very early generations, suggesting its potential as a seeding mechanism or the suitability of MO-PSO variants \cite{Lalwani_et_al:2013,Nebro_et_al:2009,Rodrigues_et_al:2016} for MORL problems.
    
    \item \textbf{Divergent performance across MORL instances}: MOEAs show highly variable performance depending on the specific MORL instance, highlighting the utility of the MORL benchmark in uncovering both strengths and limitations of current MOEAs.
    \item \textbf{Uniform performance on certain problems}: For some problems, such as \texttt{mo-humanoids-v4}, all algorithms exhibit similar performance. This may suggest that these problems are either equally challenging or equally accessible to diverse evolutionary approaches. An alternative hypothesis is that these MORLs can be addressed by neural networks capable of implementing similarly effective policies, reducing differentiation between methods.
\end{itemize}

\section{Conclusions} \label{sec:CONCLU}

Benchmarking MOEAs has historically played a critical role in advancing the field, enabling the identification of algorithmic limitations and fostering improvements. Recent research emphasizes the importance of designing benchmarks that share key characteristics with real-world problems. This consideration is particularly relevant when applying MOEAs to ML challenges.


In this paper, we have proposed the use of MORL as a platform for benchmarking MOEAs. Through extensive experiments on five well-known MORL problems, supported by statistical tests, we analyzed the performance of various MOEAs using several PF metrics and explored the impact of altering instance parameters. 


Several avenues for future research arise from this work:

\begin{itemize}
    \item \textbf{Behavior of PS solutions}: While our analysis focused on PF characteristics, deeper scrutiny is needed to analyze  PS solutions and the behavior of agents implementing policies based on these neural networks in artificial environments. Similarly, an important question is to explore the extent to which MOEAs' results translate to real-world robotic control tasks.  

    \item \textbf{Hyperparameter selection and number of repetitions}: The choice of the neural network hyperparameter (e.g., neural network architecture) and the MOEAs parameters can impact the behavior of the algorithms. An exploration of the impact of these hyperparameters could help to understand the stability and robustness of MOEAs for MORL. Similarly, the number of trials of each MOEA configuration should be increased to strengthen the statistical comparison among the algorithms. 
        
    \item \textbf{Model transferability}: Investigate the transferability of learned models across different experimental conditions (e.g., solutions trained using a few episodes and evaluated in scenarios with a larger number of episodes).
    
    \item \textbf{Evaluating MO-PSOs}: Given the promising results of PSO for scalarized problems, MO-PSOs should be extensively evaluated within the MORL framework.
    
    \item \textbf{Expanding algorithm coverage}: Include additional MOEAs and hybrid methods to further validate the versatility and challenges of the MORL benchmark and validate state-of-the-art components for each challenge presented in MORL. Moreover, we plan to perform an analysis of the advantages and disadvantages of aiming for a set of solutions compared to a scalarization in the MORL context.

    \item \textbf{Scaling in number of decision variables and objectives}: It is well known that the number of decision variables and objectives significantly affects the performance of a MOEA. Thus, we plan to perform a scalability study to further analyze the behavior of the algorithms in MORL problems.
    
    \item \textbf{Theoretical analysis of MOEAs behavior for MORL}: While the selection of a problem benchmark and the empirical analysis of MOEAs on these problems is a needed  first step, theoretical approaches that serve to characterize the MORL features and their influence on  the behavior of MOEAs for this type of problem, are needed. Recent research on theoretical analysis of MOEAs \cite{Antipov_and_Doerr:2024, Wietheger_and_Doerr:2024} could offer clues on feasible approaches to this question.  
    
\end{itemize}

By addressing these directions, we hope to advancing the application of MOEAs to increasingly realistic ML problems.

\section{Acknowledgements}

Roberto Santana acknowledges support from the Spanish Ministry of Science, Innovation and Universities (Projects ID2023-149195NB-I00 and PID2022-137442NB-I00), and the Basque Government (Grant Nos. KK-2023/00012, KK-2024/00030, and IT1504-22). Also, Carlos Hernández acknowledges the DGTIC-UNAM project IA102025.

\bibliographystyle{icml2024}
\bibliography{emo_rl}

\newpage
\appendix

\section{Explanation of the MORL problems used in the paper}

\texttt{mo-hopper-v4}: This problem involves controlling a one-legged robot (hopper) to move forward as fast as possible while maintaining balance. The agent must manage the trade-off between speed and stability, ensuring the hopper doesn't fall while maximizing forward movement.

\texttt{mo-halfcheetah-v4}: In this task, a robotic "half-cheetah" must learn to run efficiently. The agent controls a two-dimensional bipedal robot and must balance speed, energy consumption, and stability, ensuring that the cheetah remains upright and minimizes energy waste while maximizing forward velocity.

\texttt{mo-walker2d-v4}: This variant focuses on a two-legged robot (walker) that must learn to walk or run efficiently. The agent is tasked with moving forward while balancing speed, energy efficiency, and avoiding falls, making sure the walker stays upright over time.

\texttt{mo-ant-v4}: This problem involves controlling a four-legged robotic ant. The agent must navigate a three-dimensional space, optimizing for speed, energy efficiency, and stability. The complexity of movement across multiple joints and directions requires the agent to learn coordinated and energy-efficient strategies to avoid tipping over.

\texttt{mo-humanoid-v4}: The most complex of the set, this problem requires controlling a humanoid robot with many degrees of freedom. The agent must optimize for walking or running speed while maintaining balance and minimizing energy usage. The humanoid's complexity makes it challenging to learn efficient, coordinated movement that achieves high rewards across multiple objectives.

\section{Behavior in terms of the HV for $n\_episodes =5$ and $ngen=25$}

\begin{figure}[]
     \begin{center}  
      \includegraphics[width=7.5cm]{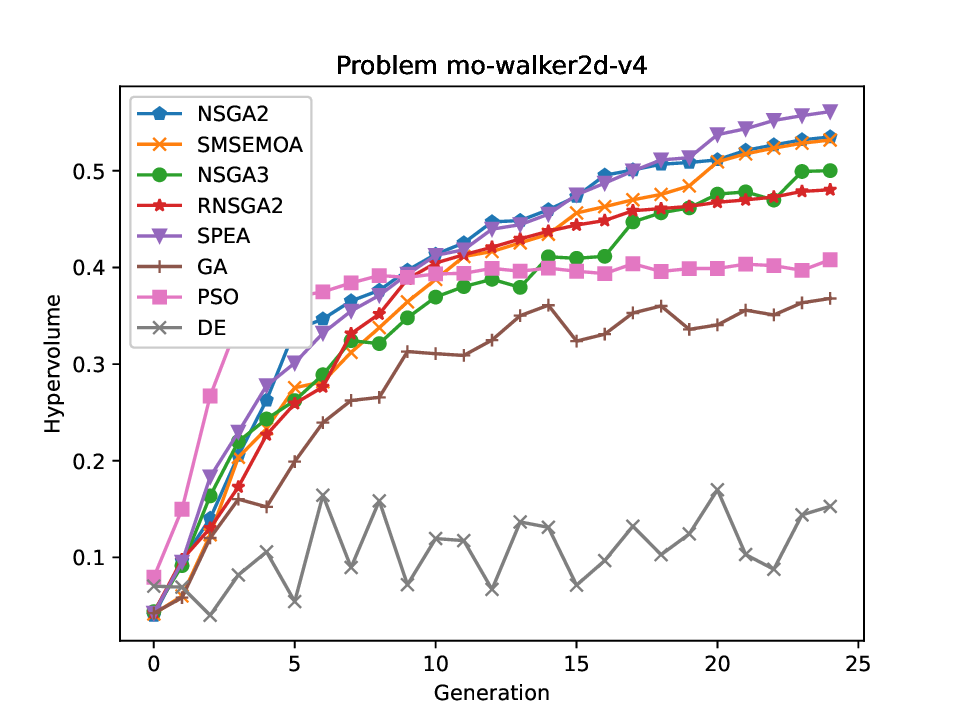}   
\caption{Evolution of the HV for problem  \texttt{mo-walker2d-v4}, $n\_episodes=5$.}
  \label{fig:RESULTS_CM_2_NEP_5}
 \end{center}
   \end{figure}

\begin{figure}[]
     \begin{center}  
      \includegraphics[width=7.5cm]{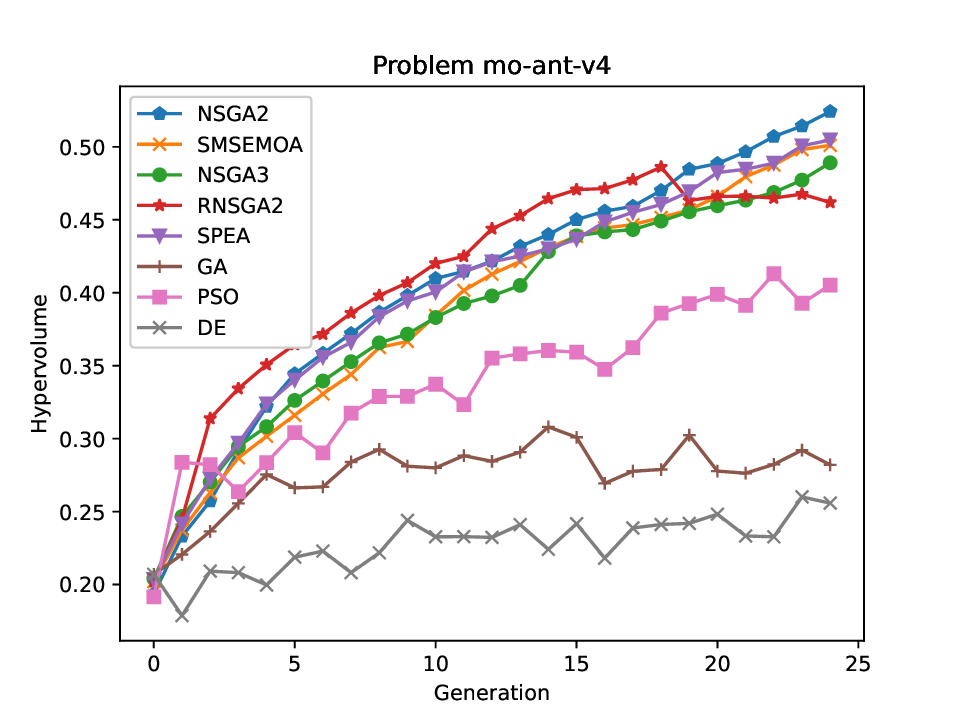}   
\caption{Evolution of the HV for problem  \texttt{mo-ant-v4}, $n\_episodes=5$.}
  \label{fig:RESULTS_CM_3_NEP_5}
 \end{center}
   \end{figure}

\begin{figure}[]
     \begin{center}
      \includegraphics[width=7.5cm]{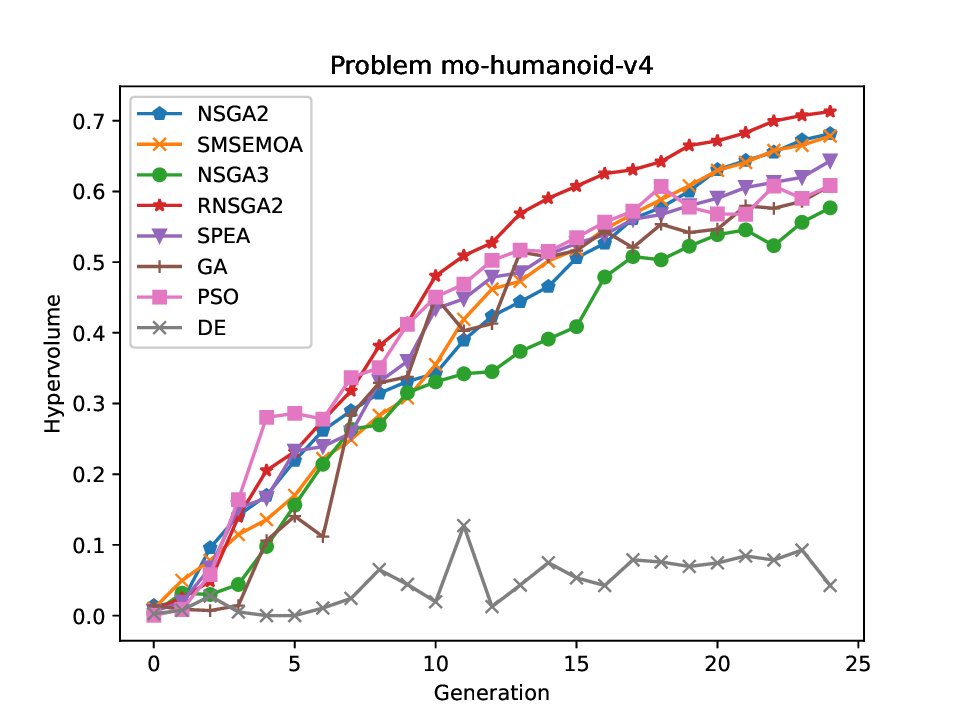}
\caption{Evolution of the HV for problem  \texttt{mo-humanoid-v4}, $n\_episodes=5$.}
\label{fig:RESULTS_CM_4_NEP_5}
     \end{center}
   \end{figure}

Figures~\ref{fig:RESULTS_CM_2_NEP_5}, ~\ref{fig:RESULTS_CM_3_NEP_5}, and~\ref{fig:RESULTS_CM_4_NEP_5} respectively show the behavior in terms of the HV for problems \texttt{mo-walker2d-v4}, \texttt{mo-ant-v4} and~\texttt{mo-humanoid-v4} when $n\_episodes =5$ and $ngen=25$
  
\newpage
\section{PFs found for the MORL problems}

\begin{figure}[htbp]
     \begin{center}
      \includegraphics[width=7.5cm]{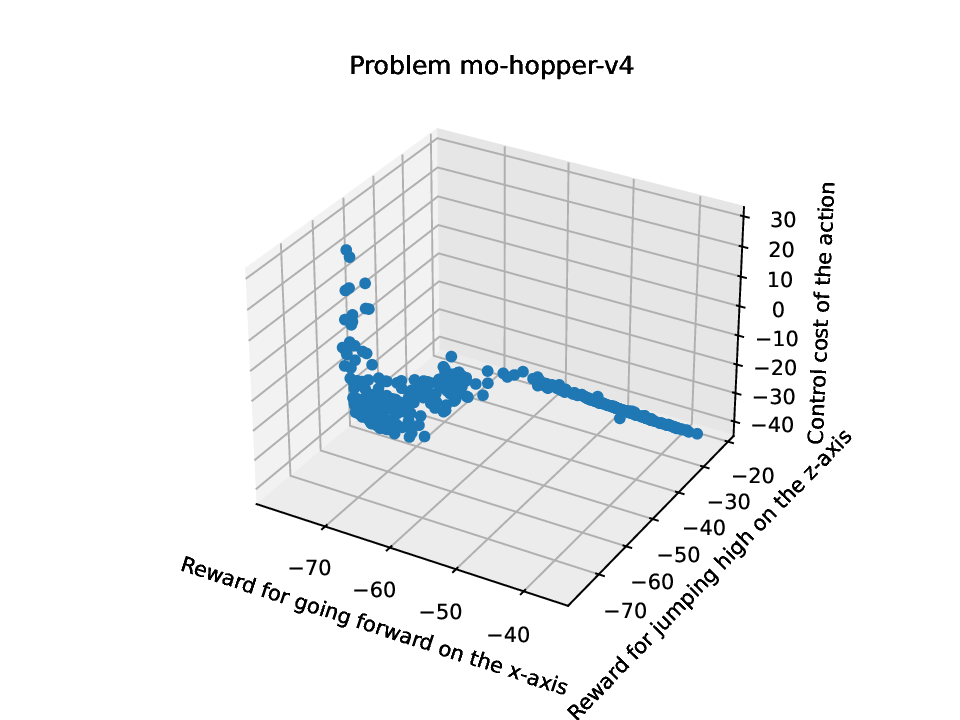}
   \caption{Pareto front for problem  \texttt{mo-hopper-v4}, $n\_episodes=5$.}    
   \label{fig:PF_CM_0_NEP_5}
 \end{center}
    \end{figure}

\begin{figure}[]
     \begin{center}
      \includegraphics[width=7.5cm]{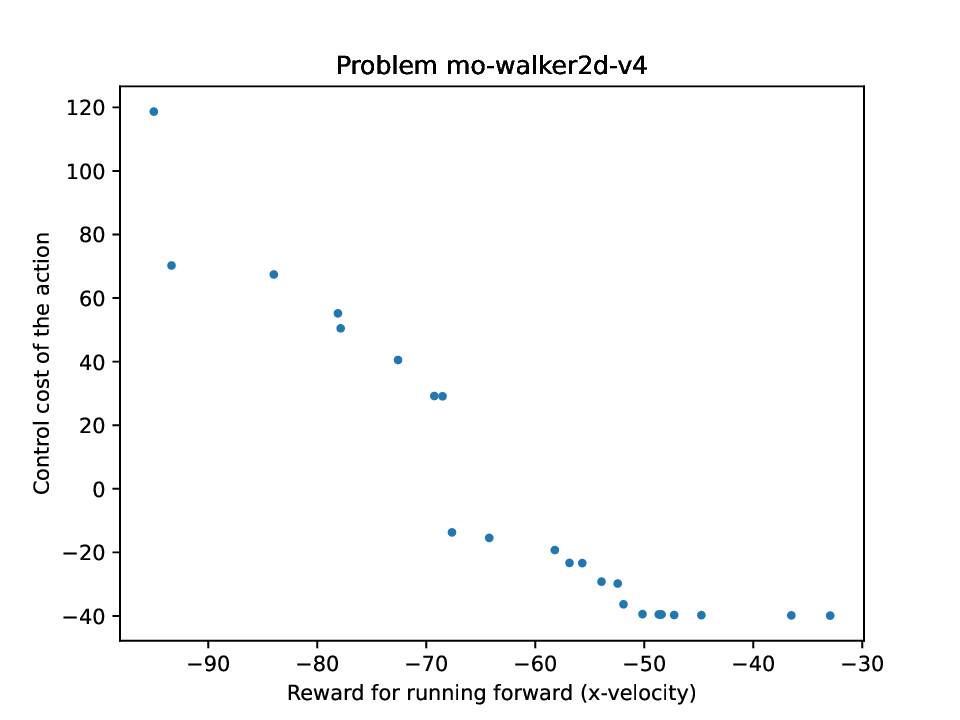}
   \caption{Pareto front for problem  \texttt{mo-walker2d-v4}, $n\_episodes=5$.}   
   \label{fig:PF_CM_2_NEP_5}
 \end{center}
   \end{figure}

\begin{figure}[]
     \begin{center}
      \includegraphics[width=7.5cm]{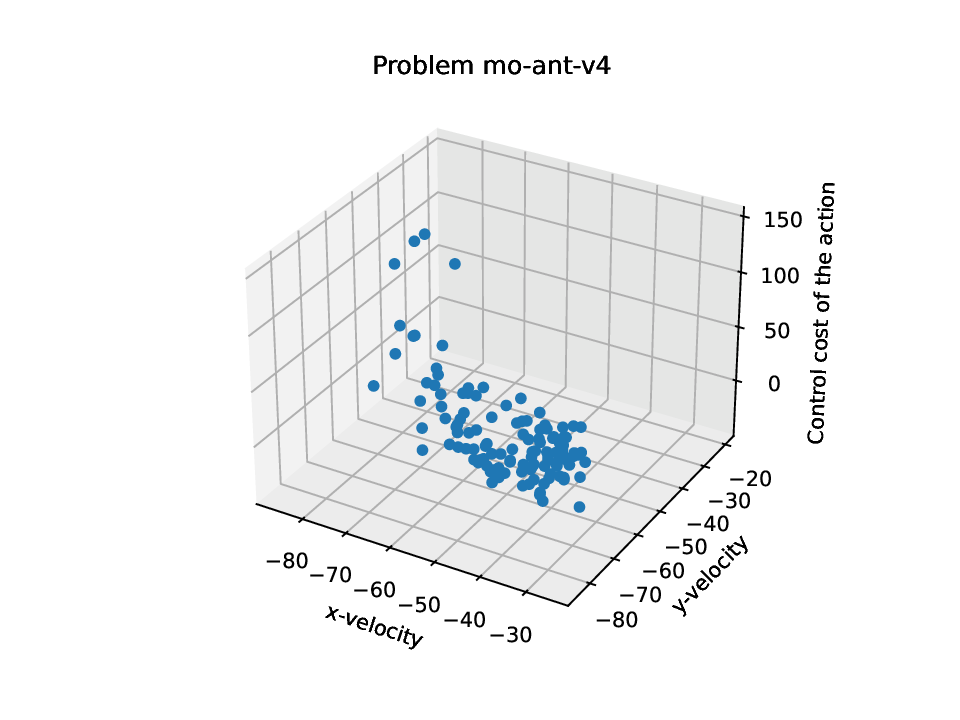}
   \caption{Pareto front for problem  \texttt{mo-ant-v4}, $n\_episodes=5$.}     
  \label{fig:PF_CM_3_NEP_5}
 \end{center}
   \end{figure}

Figures~\ref{fig:PF_CM_0_NEP_5}, ~\ref{fig:PF_CM_2_NEP_5}, and~\ref{fig:PF_CM_3_NEP_5} respectively show the PFs found for problems \texttt{mo-hopper-v4}, \texttt{mo-walker2d-v4}, and \texttt{mo-ant-v4}  when   $n\_episodes =5$ and $ngen=25$

\newpage
\section{Behavior in terms of the GD and IGD metrics for $n\_episodes =5$ and $ngen=2$}

\begin{figure}[htbp]
     \begin{center}
      \includegraphics[width=7.5cm]{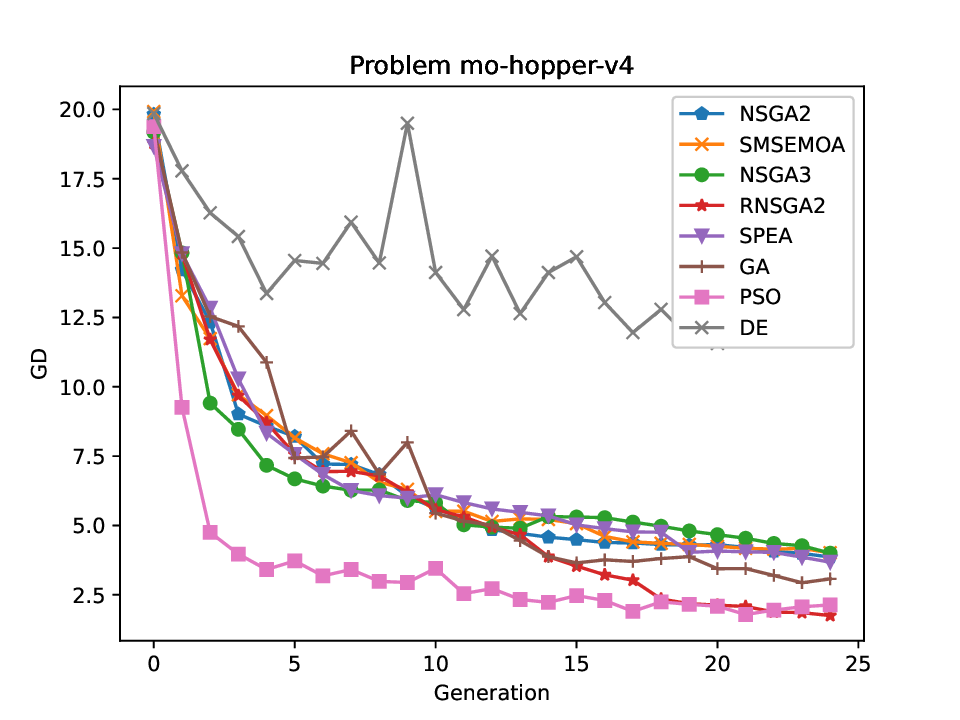}
   \caption{Evolution of the GD metric for problem  \texttt{mo-hopper-v4}, $n\_episodes=5$.}    \label{fig:GD_CM_0_NEP_5}
 \end{center}
   \end{figure}

\begin{figure}[htbp]
     \begin{center}
      \includegraphics[width=7.5cm]{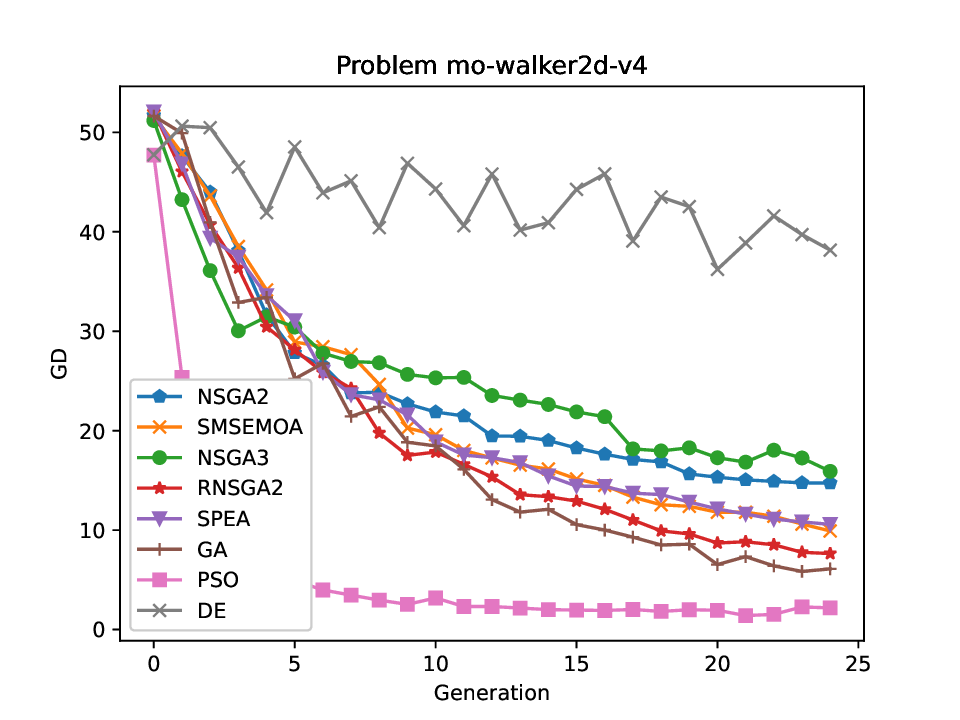}
   \caption{Evolution of the GD metric for problem  \texttt{mo-walker2d-v4}, $n\_episodes=5$.}   \label{fig:GD_CM_2_NEP_5}
 \end{center}
   \end{figure}

\begin{figure}[htbp]
     \begin{center}
      \includegraphics[width=7.5cm]{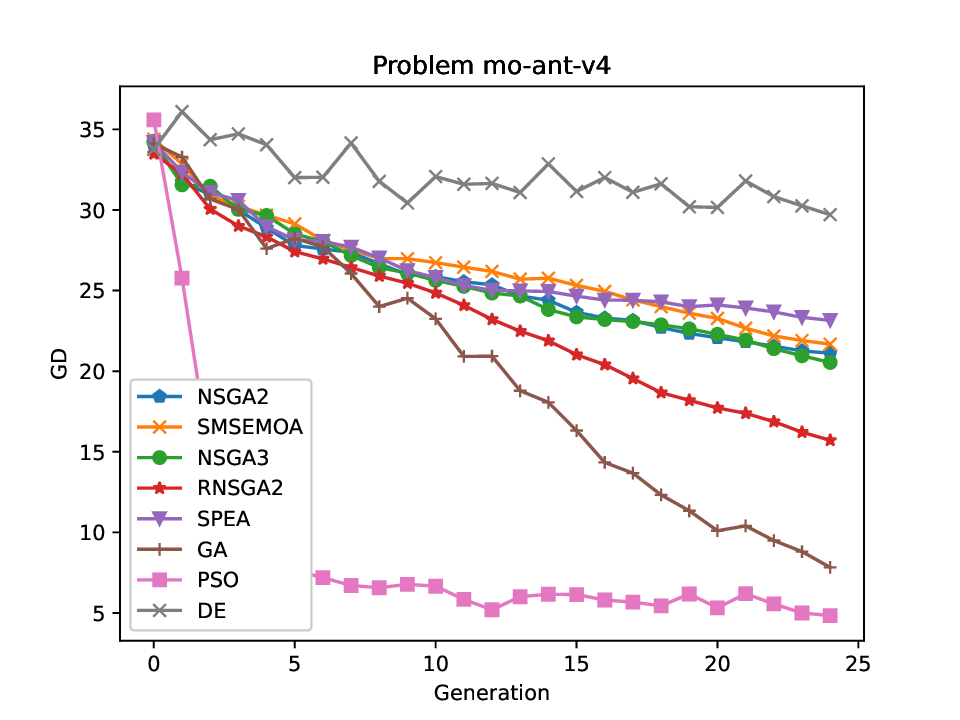}
   \caption{Evolution of the GD metric for problem  \texttt{mo-ant-v4}, $n\_episodes=5$.}     
  \label{fig:GD_CM_3_NEP_5}
 \end{center}
   \end{figure}

\begin{figure}[htbp]
     \begin{center}
      \includegraphics[width=7.5cm]{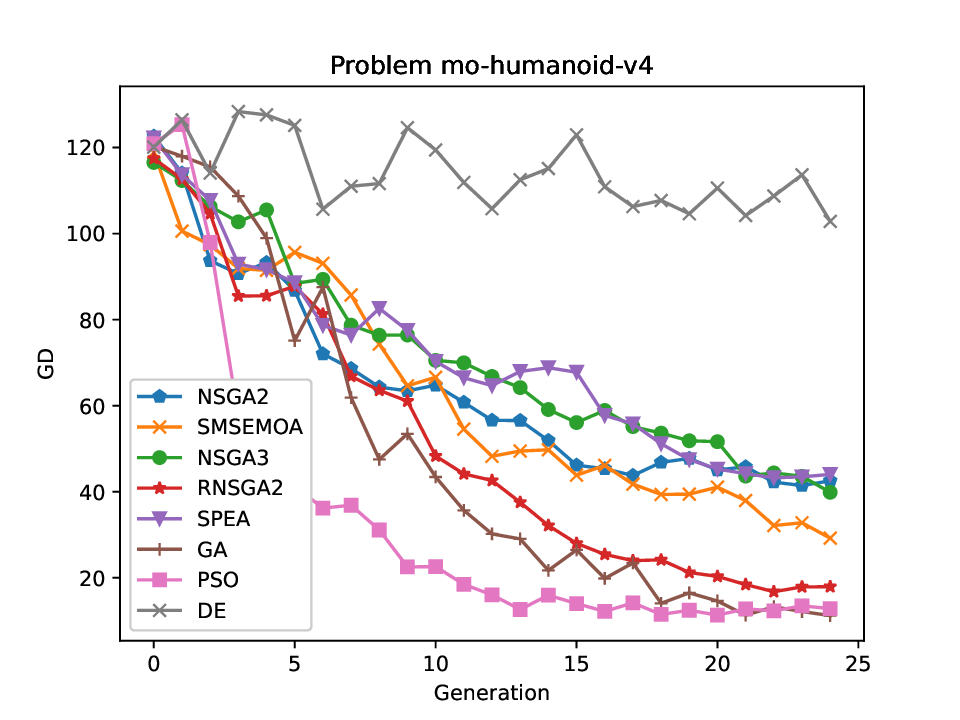}
   \caption{Evolution of the GD metric for problem  \texttt{mo-humanoid-v4}, $n\_episodes=5$.}        \label{fig:GD_CM_4_NEP_5}
 \end{center}
   \end{figure}

\begin{figure}[htbp]
     \begin{center}
      \includegraphics[width=7.5cm]{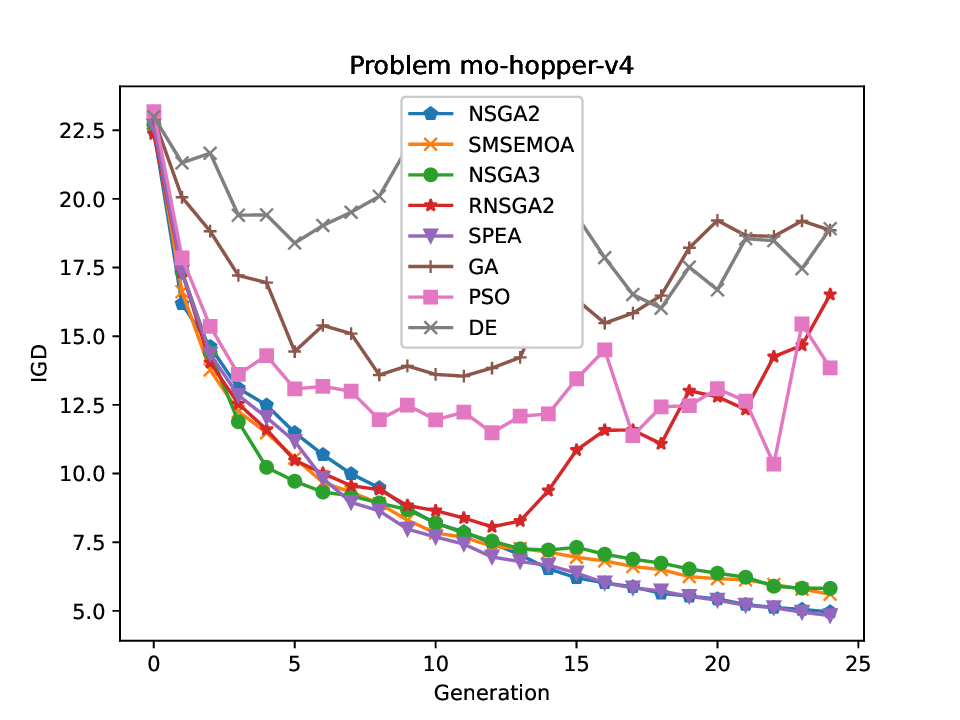}
   \caption{Evolution of the IGD metric for problem  \texttt{mo-hopper-v4}, $n\_episodes=5$.}    \label{fig:IGD_CM_0_NEP_5}
 \end{center}
   \end{figure}

\begin{figure}[htbp]
     \begin{center}
      \includegraphics[width=7.5cm]{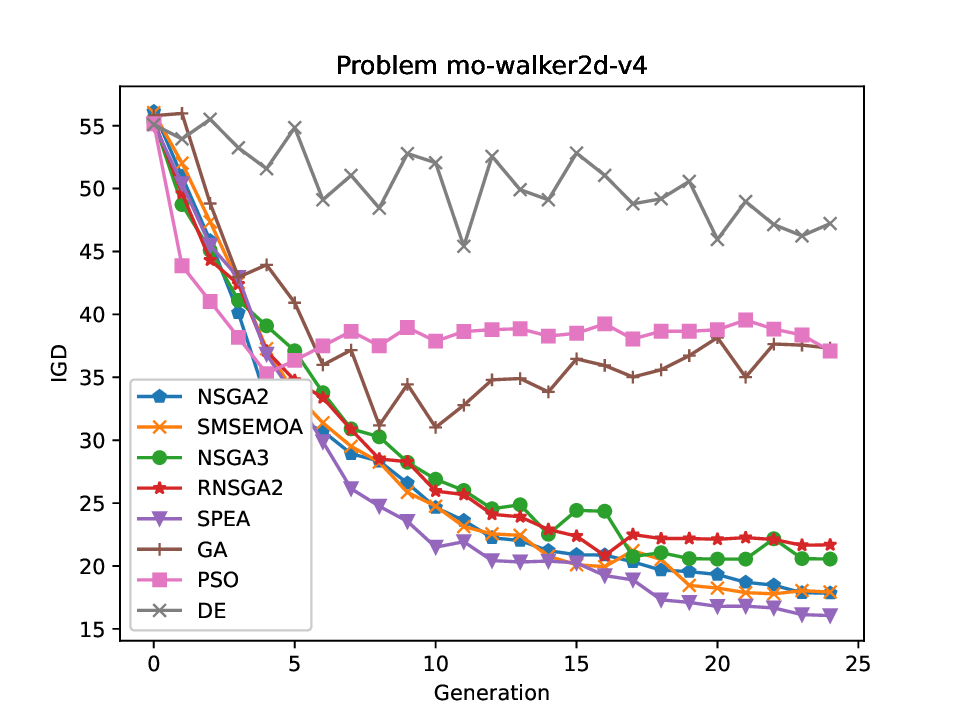}
   \caption{Evolution of the IGD metric for problem  \texttt{mo-walker2d-v4}, $n\_episodes=5$.}   \label{fig:IGD_CM_2_NEP_5}
 \end{center}
   \end{figure}

\begin{figure}[htbp]
     \begin{center}
      \includegraphics[width=7.5cm]{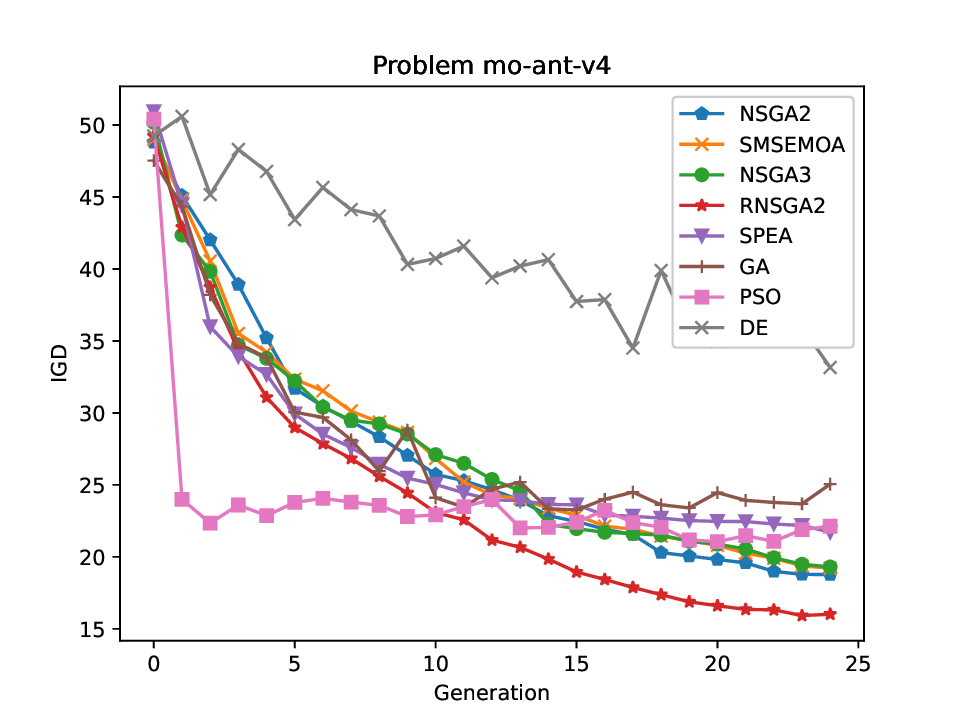}
   \caption{Evolution of the IGD metric for problem  \texttt{mo-ant-v4}, $n\_episodes=5$.}     
  \label{fig:IGD_CM_3_NEP_5}
 \end{center}
   \end{figure}

\begin{figure}[htbp]
     \begin{center}
      \includegraphics[width=7.5cm]{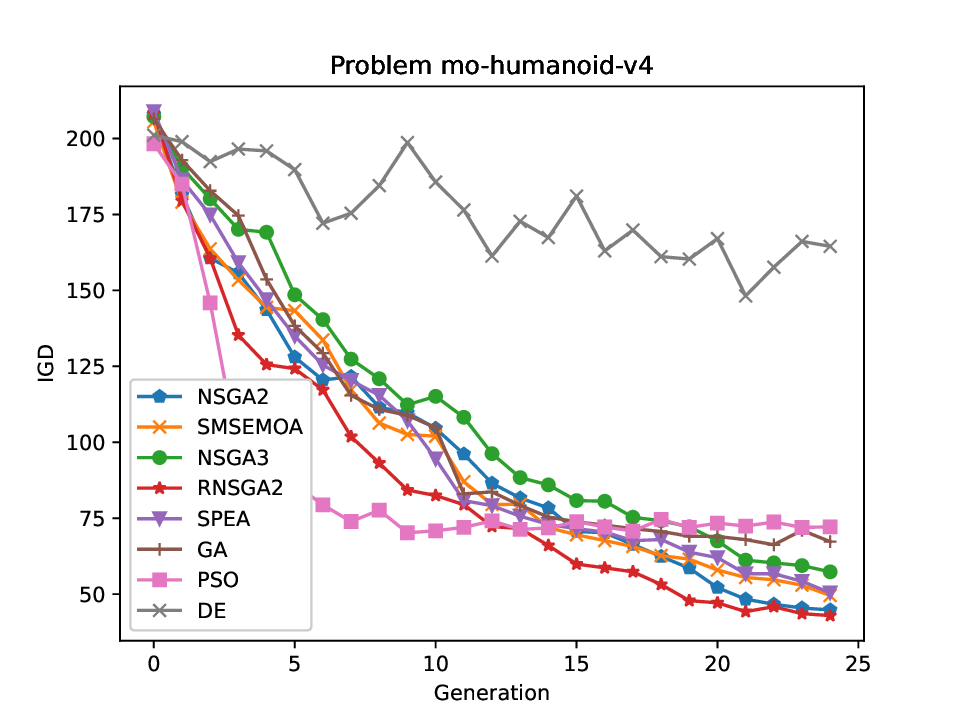}
   \caption{Evolution of the IGD metric for problem  \texttt{mo-humanoid-v4}, $n\_episodes=5$.}         
   \label{fig:IGD_CM_4_NEP_5}
 \end{center}
   \end{figure}

Figures~\ref{fig:GD_CM_0_NEP_5}~-\ref{fig:GD_CM_4_NEP_5} show the evolution of the GD metric for all problems. Figures~\ref{fig:IGD_CM_0_NEP_5}~-\ref{fig:IGD_CM_4_NEP_5} show the evolution of the IGD metric for all problems.

\newpage
\section{Evolution of the scalarized values}

\begin{figure}[htbp]
     \begin{center}
      \includegraphics[width=7.5cm]{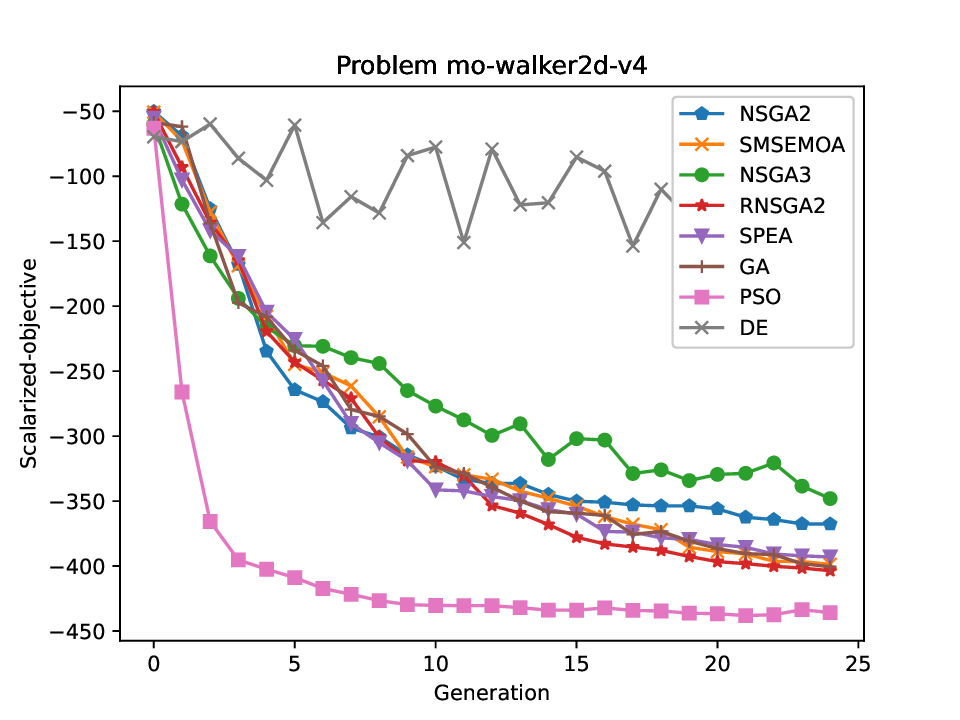}
   \caption{Evolution of the scalarized objective for problem  \texttt{mo-walker2d-v4}, $n\_episodes=5$.}   \label{fig:SCA_CM_2_NEP_5}
 \end{center}
   \end{figure}

\begin{figure}[htbp]
     \begin{center}
      \includegraphics[width=7.5cm]{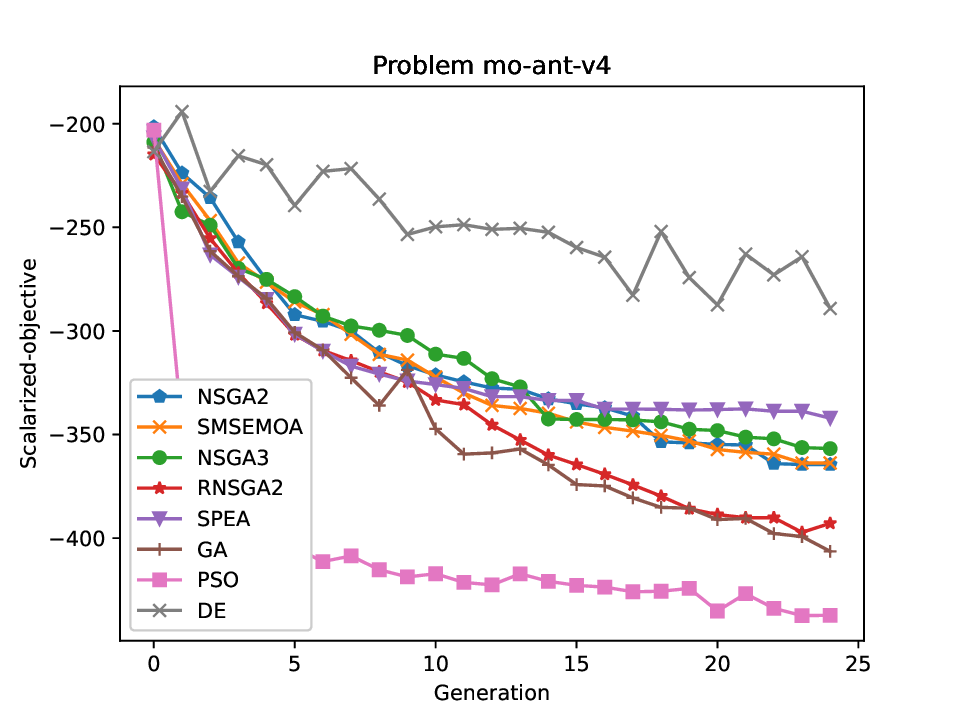}
   \caption{Evolution of the scalarized objective for problem  \texttt{mo-ant-v4}, $n\_episodes=5$.}     
  \label{fig:SCA_CM_3_NEP_5}
 \end{center}
   \end{figure}

\begin{figure}[htbp]
     \begin{center}
      \includegraphics[width=7.5cm]{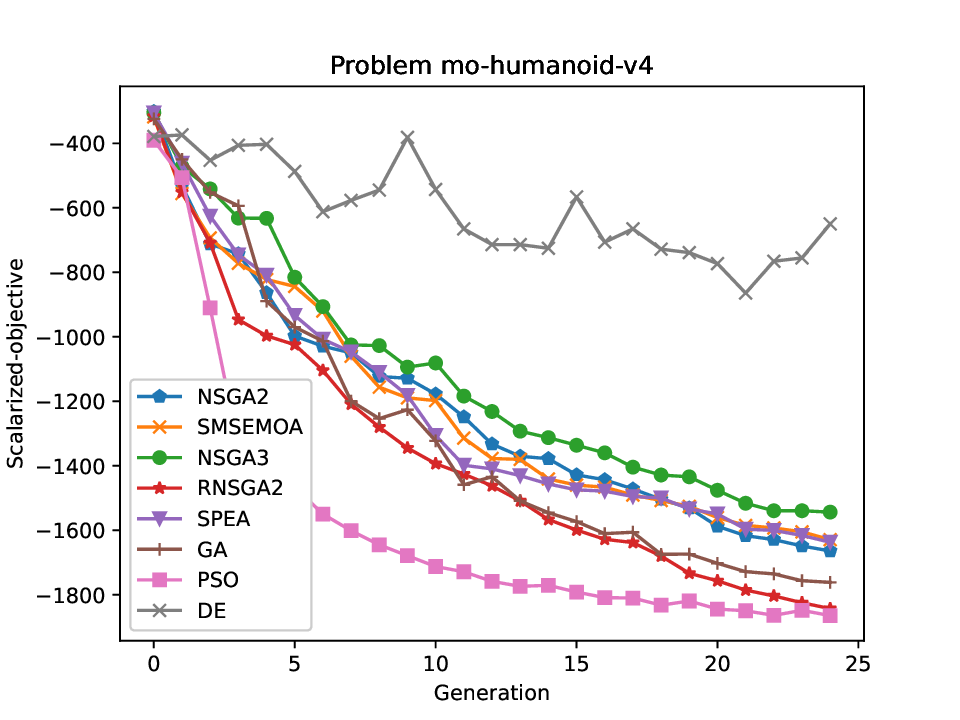}
   \caption{Evolution of the scalarized objective for problem  \texttt{mo-humanoid-v4}, $n\_episodes=5$.}         
   \label{fig:SCA_CM_4_NEP_5}
 \end{center}
   \end{figure}

Figures~\ref{fig:SCA_CM_2_NEP_5}-~\ref{fig:SCA_CM_4_NEP_5} show the evolution of the scalarized values for all problems.


   
\newpage
\section{Results for a higher number of episodes ($n\_episodes =10$)}

\begin{figure}[htbp]
     \begin{center}
     \includegraphics[width=7.5cm]{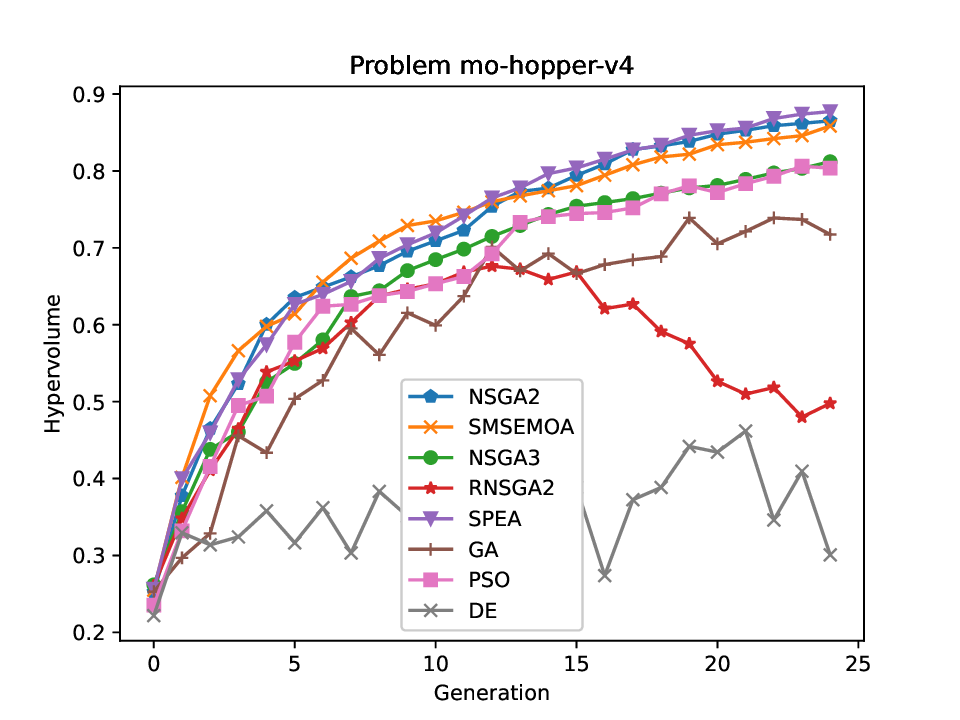}    
   \caption{Evolution of the HV for problem  \texttt{mo-hopper-v4}, $n\_episodes=10$.}       
  \label{fig:RESULTS_CM_0_NEP_10}
 \end{center}
   \end{figure}

\begin{figure}[htbp]
     \begin{center}
    \includegraphics[width=7.5cm]{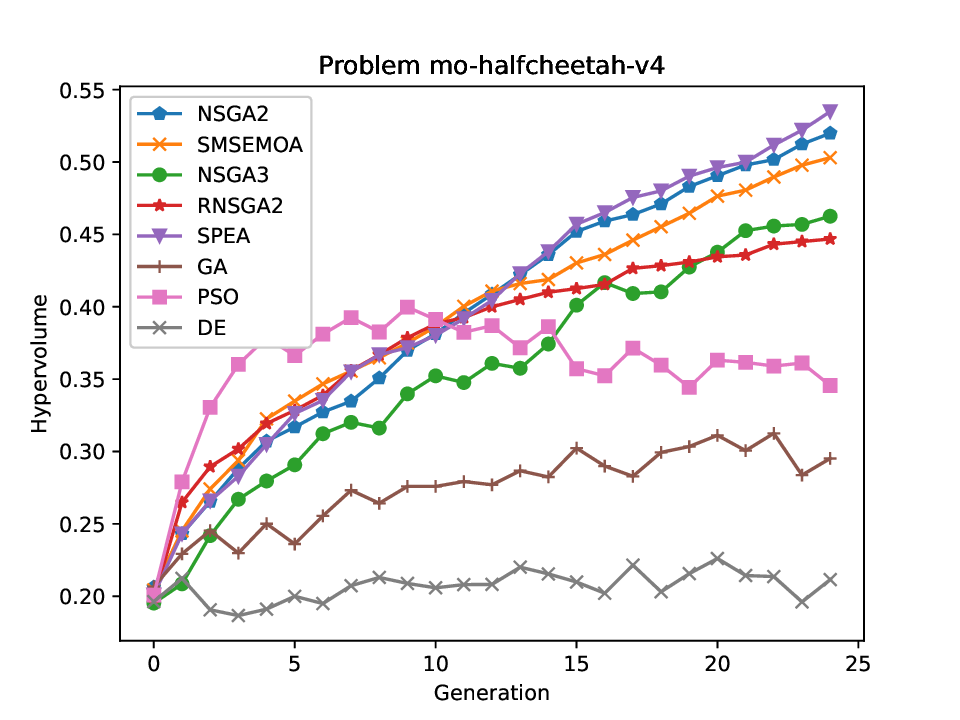}
   \caption{Evolution of the HV for problem  \texttt{mo-halfcheetah-v4}, $n\_episodes=10$.}       
  \label{fig:RESULTS_CM_1_NEP_10}
 \end{center}
   \end{figure}

\begin{figure}[htbp]
     \begin{center}
      \includegraphics[width=7.5cm]{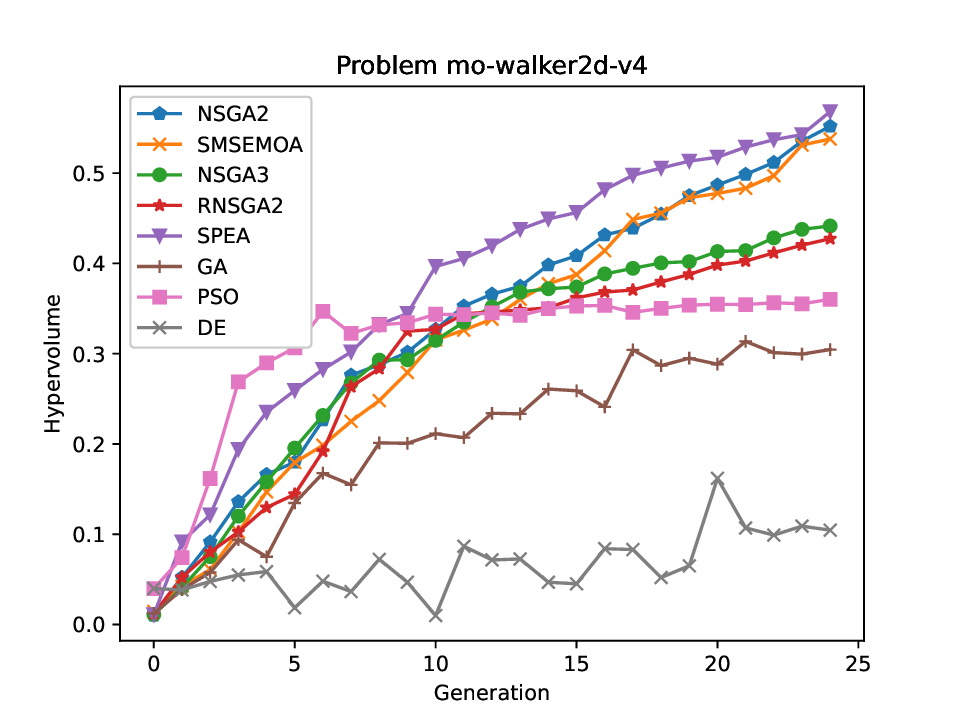}
   \caption{Evolution of the HV for problem  \texttt{mo-walker2d-v4}, $n\_episodes=10$.}       
  \label{fig:RESULTS_CM_2_NEP_10}
 \end{center}
   \end{figure}

\begin{figure}[htbp]
     \begin{center}
      \includegraphics[width=7.5cm]{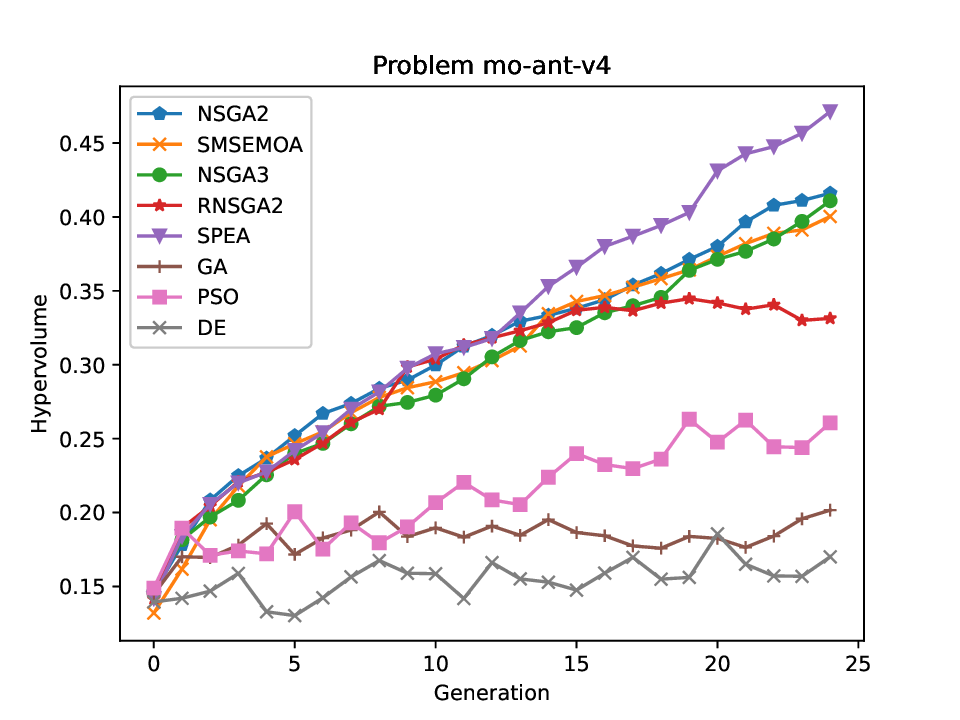}
   \caption{Evolution of the HV for problem  \texttt{mo-ant-v4}, $n\_episodes=10$.}       
  \label{fig:RESULTS_CM_3_NEP_10}
 \end{center}
   \end{figure}

\begin{figure}[htbp]
     \begin{center}
           \includegraphics[width=7.5cm]{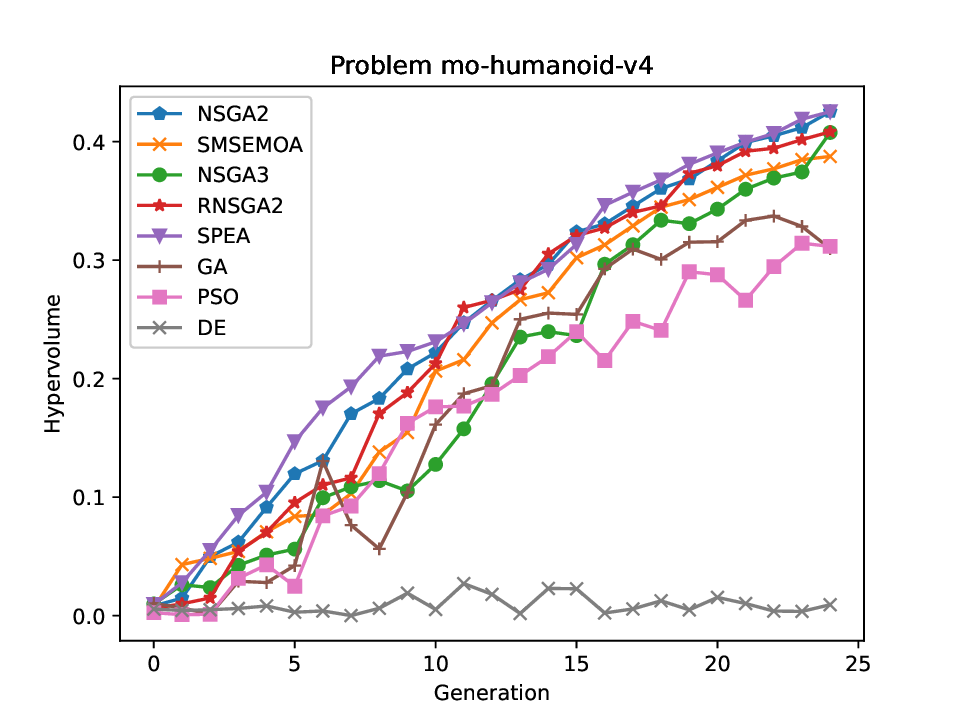}

\caption{Evolution of the HV for problem  \texttt{mo-humanoid-v4}, $n\_episodes=10$.}
\label{fig:RESULTS_CM_4_NEP_10}
 \end{center}
   \end{figure}

 Figures~\ref{fig:RESULTS_CM_0_NEP_10}-~\ref{fig:RESULTS_CM_4_NEP_10} show the evolution of the HV for all problems when $n\_episodes =10$. 


\newpage
\section{Results for a model with a higher number of parameters}

\begin{figure}[htbp]
     \begin{center}
      \includegraphics[width=7.5cm]{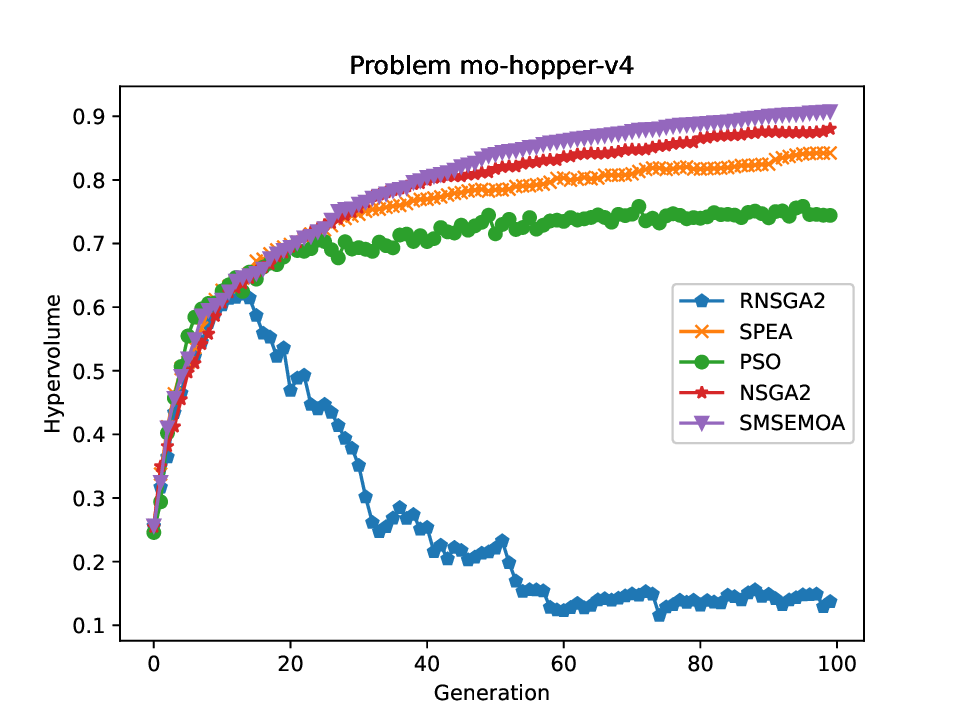}
   \caption{Evolution of the HV for problem  \texttt{mo-hopper-v4}, $n\_episodes=10$, $n\_layer\_2 = 10$.}       
  \label{fig:RESULTS_CM_0_NEP_10_Gen_100}
 \end{center}
   \end{figure}

\begin{figure}[htbp]
     \begin{center}
      \includegraphics[width=7.5cm]{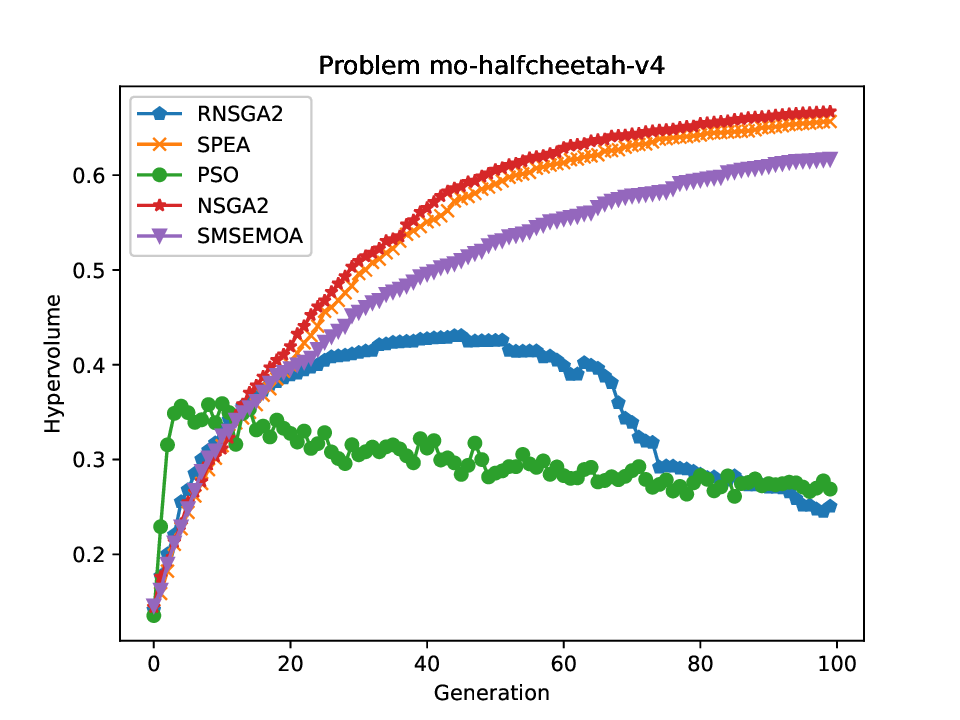}
   \caption{Evolution of the HV for problem  \texttt{mo-halfcheetah-v4}, $n\_episodes=10$, $n\_layer\_2 = 10$.}       
  \label{fig:RESULTS_CM_1_NEP_10_Gen_100}
 \end{center}
   \end{figure}

\begin{figure}[htbp]
     \begin{center}  
      \includegraphics[width=7.5cm]{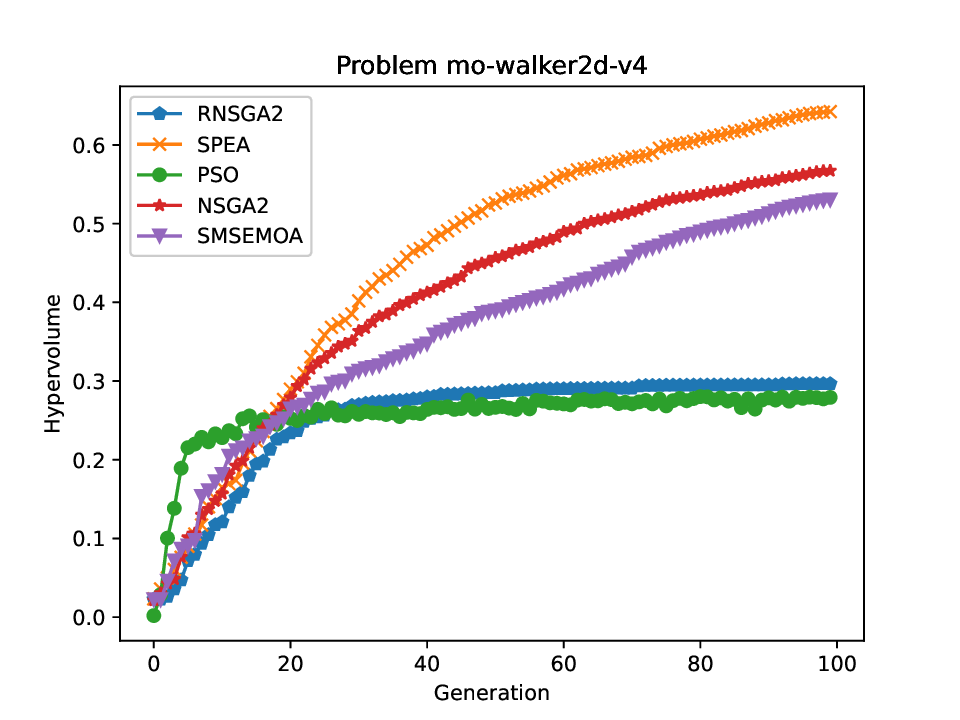}
\caption{Evolution of the HV for problem  \texttt{mo-walker2d-v4}, $n\_episodes=10$, $n\_layer\_2 = 10$.}
  \label{fig:RESULTS_CM_2_NEP_10_Gen_100}
 \end{center}
   \end{figure}

\begin{figure}[htbp]
     \begin{center}  
      \includegraphics[width=7.5cm]{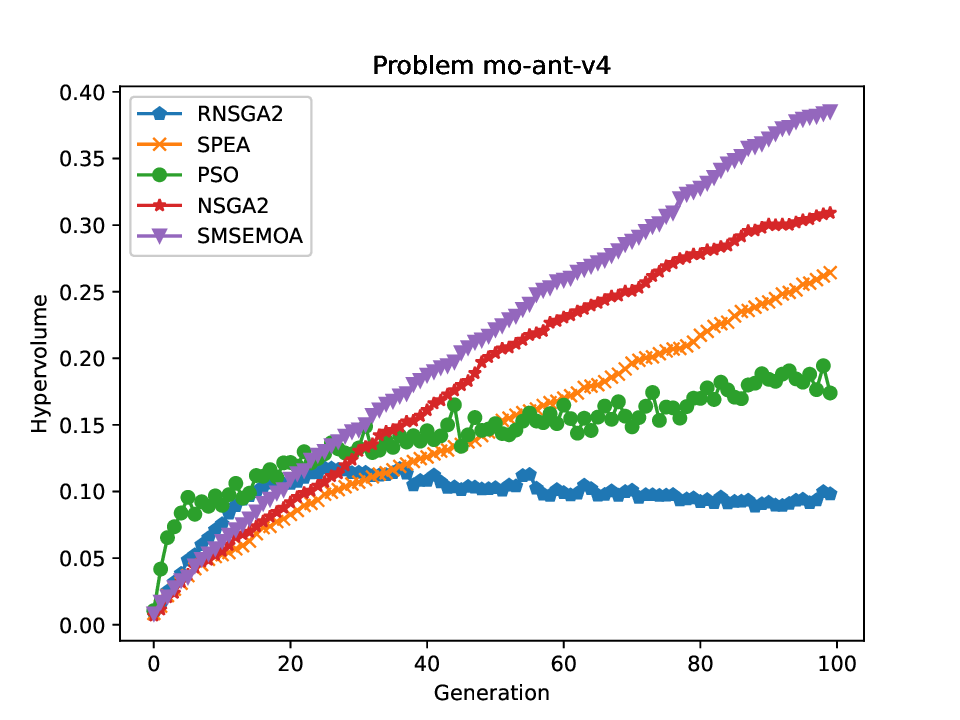}
\caption{Evolution of the HV for problem  \texttt{mo-ant-v4}, $n\_episodes=10$, $n\_layer\_2 = 10$.}
  \label{fig:RESULTS_CM_3_NEP_10_Gen_100}
 \end{center}
   \end{figure}

\begin{figure}[htbp]
     \begin{center}
      \includegraphics[width=7.5cm]{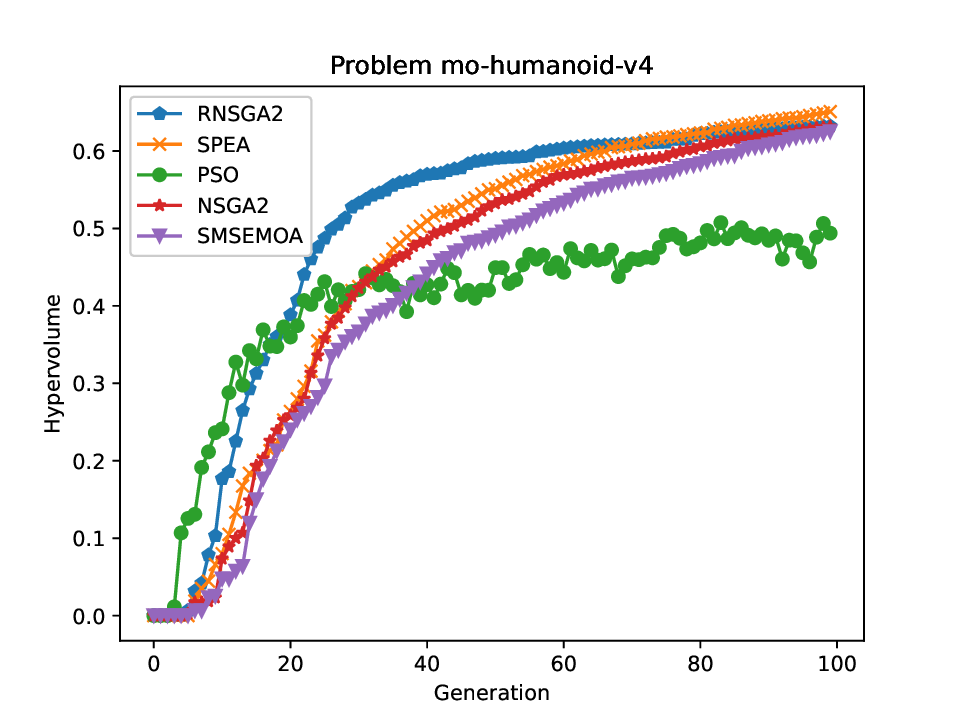}
\caption{Evolution of the HV for problem  \texttt{mo-humanoid-v4}, $n\_episodes=10$, $n\_layer\_2 = 10$.}
\label{fig:RESULTS_CM_4_NEP_10_Gen_100}
 \end{center}
   \end{figure}

 Figures~\ref{fig:RESULTS_CM_0_NEP_10_Gen_100}-~\ref{fig:RESULTS_CM_4_NEP_10_Gen_100} show the evolution of the HV for all problems when  $n\_layer2 =10$ and $n\_episodes =10$. 
 
\begin{figure}[htbp]
     \begin{center}
      \includegraphics[width=7.5cm]{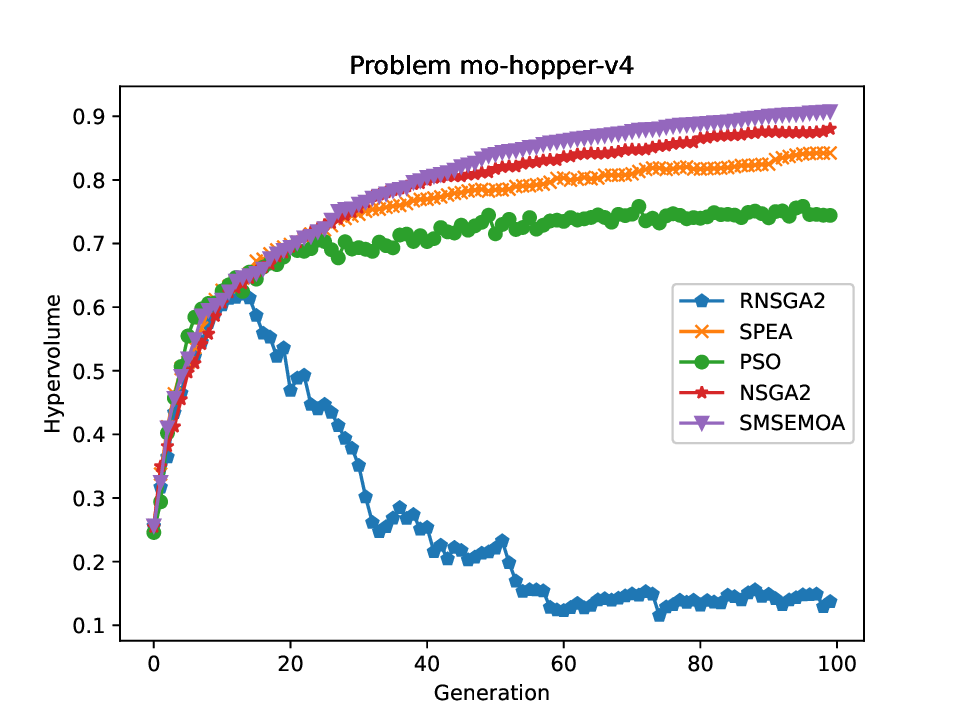}
   \caption{Evolution of the HV for problem  \texttt{mo-hopper-v4}, $n\_episodes=30$ and $n\_layer\_2=10$.}      
  \label{fig:RESULTS_CM_1_NEP_30}
 \end{center}
   \end{figure}

 Figure~\ref{fig:RESULTS_CM_1_NEP_30} shows the evolution of the HV for problem  \texttt{mo-hopper-v4} when  $n\_layer2 =10$ and $n\_episodes =30$.

\end{document}